\documentclass[runningheads]{llncs}

\usepackage{eccv}

\usepackage{eccvabbrv}

\usepackage{graphicx}
\usepackage{booktabs}

\usepackage{pgfplots}
\usepgfplotslibrary{statistics}
\pgfplotsset{compat=1.18}
\usepackage{subcaption}
\usepackage{moresize}
\usepackage{tikz}
\setkeys{Gin}{keepaspectratio}
\usepackage{amsmath}
\usepackage{amssymb}
\usepackage{booktabs}
\usepackage[dvipsnames, table]{xcolor}
\usepackage{float}
\usepackage{pgfplots}
\usepackage{tikz}
\usetikzlibrary{patterns}
\usepackage[precision=2, unit=mm]{lengthconvert}
\usepackage{diagbox}
\usepackage{enumitem}
\usepackage{tabularx}
\usepackage{multirow}
\usepackage{wrapfig}
\usepackage{cuted}
\usepackage{overpic}
\usepackage[rightcaption]{sidecap}
\usepackage{calc}
\usepackage{bbm}
\usepackage{adjustbox}
\pgfplotsset{compat=1.18}

\definecolor{mycolor1}{RGB}{90,130,213}
\definecolor{mycolor3}{RGB}{231,92,46}
\definecolor{mycolor11}{RGB}{134,168,235}
\definecolor{mycolor33}{RGB}{231,156,130}
\definecolor{mycolor2}{RGB}{230,130,46}
\definecolor{cBLUE}{RGB}{90,130,213}%
\definecolor{cBLUE1}{RGB}{90,183,214}%
\definecolor{cBLUE2}{RGB}{132,217,226}%
\definecolor{cRED}{RGB}{231,92,46}%
\definecolor{cPINK}{RGB}{200,57,170}%
\definecolor{cPINKLIGHT}{RGB}{255,209,245}%
\definecolor{cGREEN}{RGB}{80,150,80}%
\definecolor{cYELLOW}{RGB}{247,179,43}%
\definecolor{cORANGE}{RGB}{242,105,0}%
\definecolor{cGRAY}{RGB}{129,141,146}%

\newlength{\figAwidth}
\newlength{\figBwidth}

\newlength{\fHeight}
\newcommand{\faBicycle}{\includegraphics[height=\fHeight]{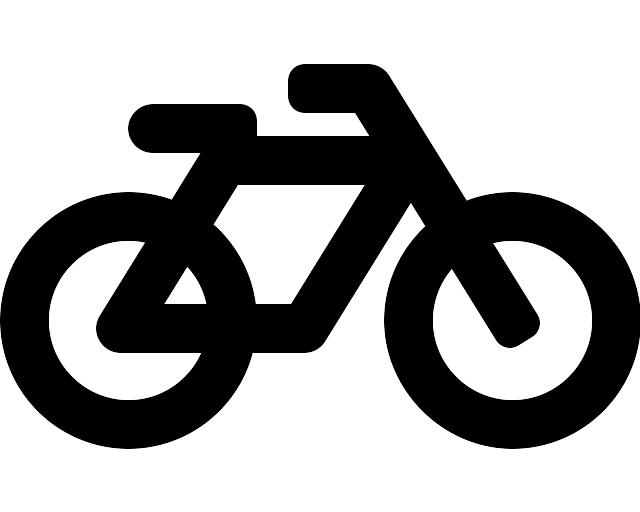}}
\newcommand{\faBus}{\includegraphics[height=\fHeight]{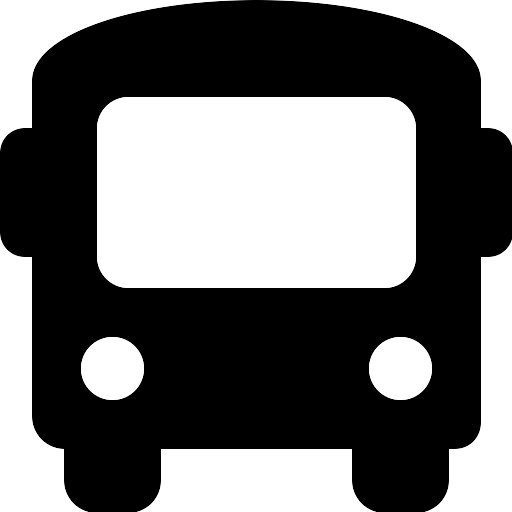}}
\newcommand{\faCar}{\includegraphics[height=\fHeight]{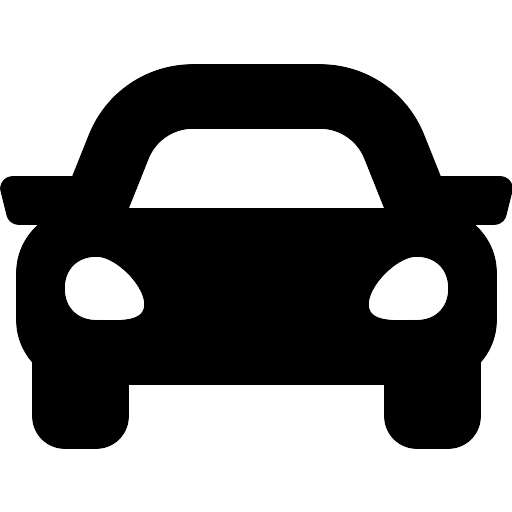}}
\newcommand{\faCat}{\includegraphics[height=\fHeight]{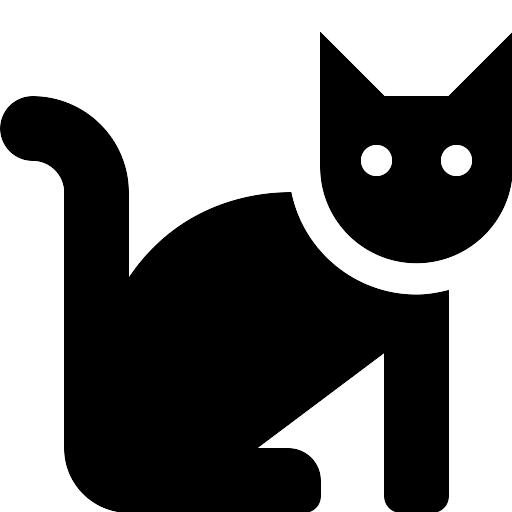}}
\newcommand{\faCow}{\includegraphics[height=\fHeight]{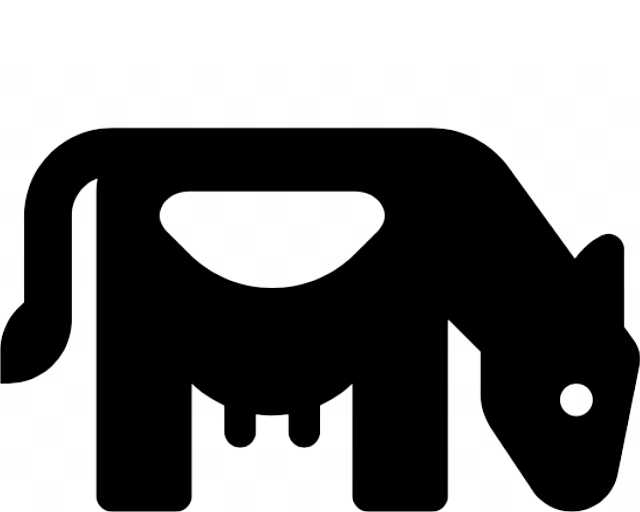}}
\newcommand{\faChair}{\includegraphics[height=\fHeight]{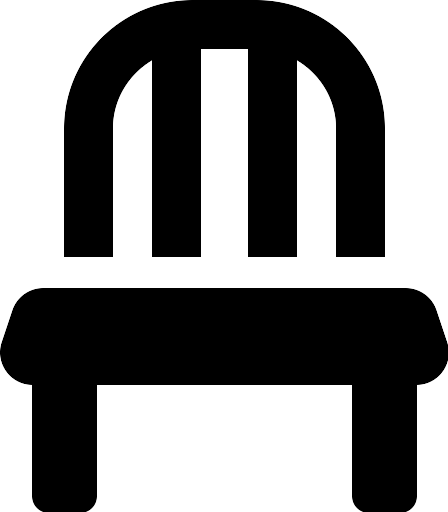}}
\newcommand{\faCouch}{\includegraphics[height=\fHeight]{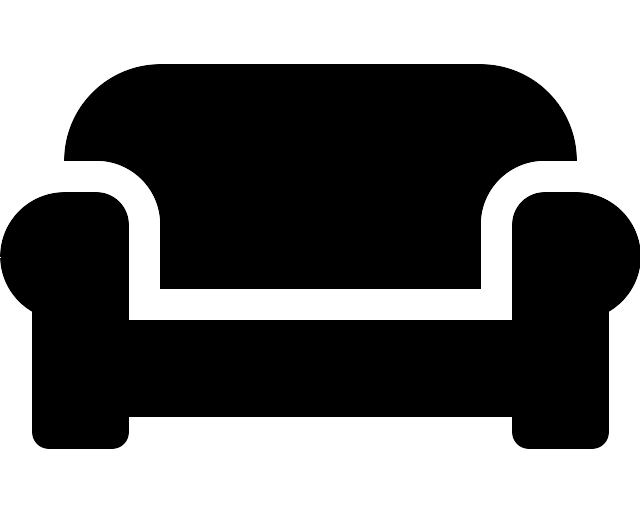}}
\newcommand{\faCrow}{\includegraphics[height=\fHeight]{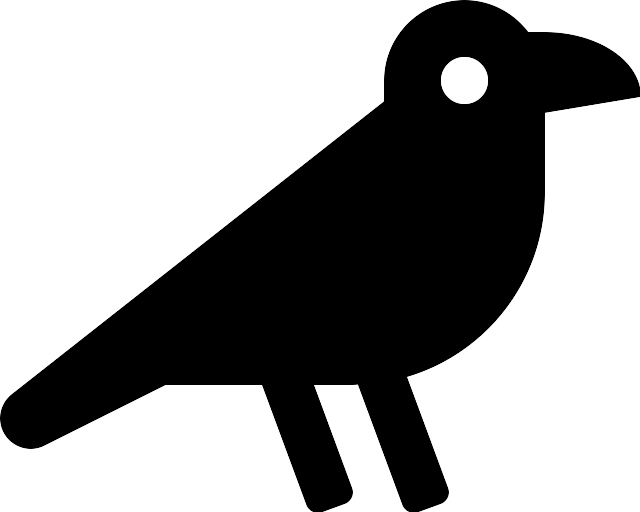}}
\newcommand{\faDog}{\includegraphics[height=\fHeight]{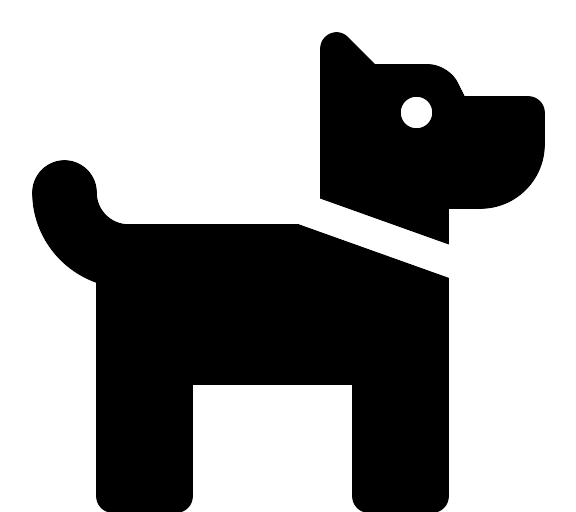}}
\newcommand{\faDinningTable}{\includegraphics[height=\fHeight]{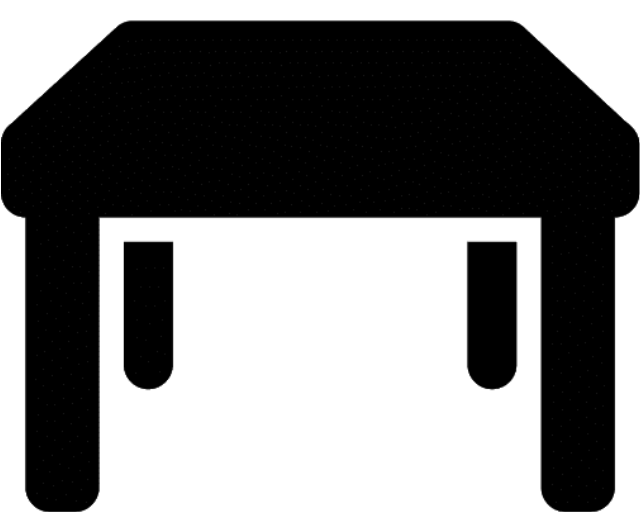}}
\newcommand{\faHorse}{\includegraphics[height=\fHeight]{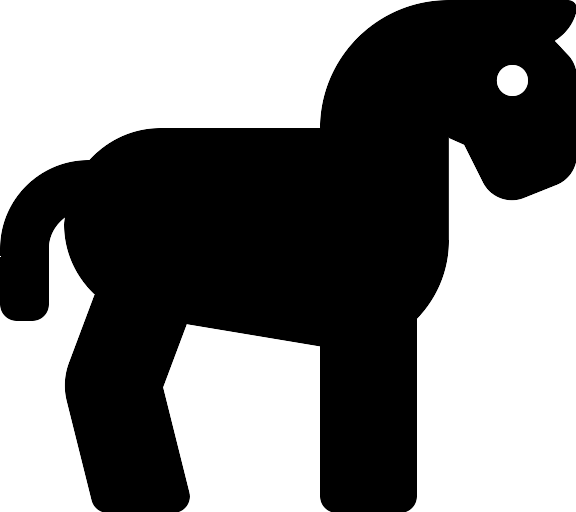}}
\newcommand{\faMotorcycle}{\includegraphics[height=\fHeight]{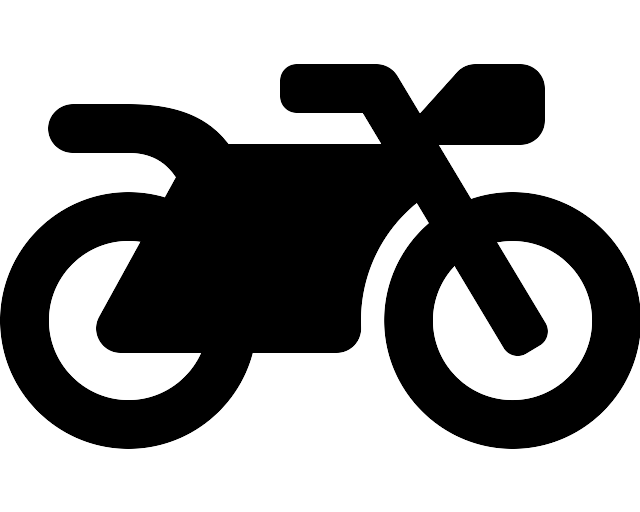}}
\newcommand{\faPlane}{\includegraphics[height=\fHeight]{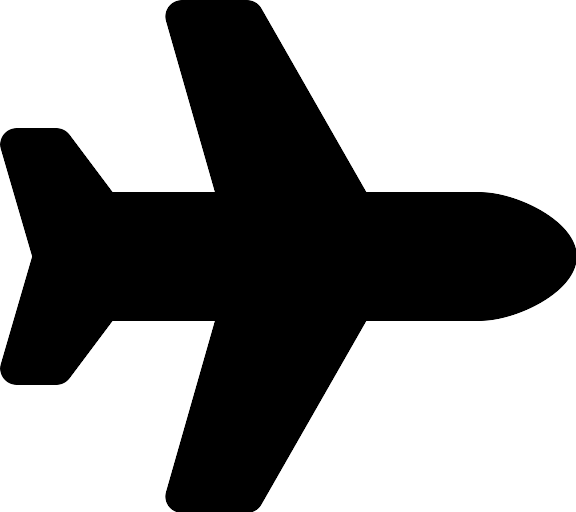}}
\newcommand{\faShip}{\includegraphics[height=\fHeight]{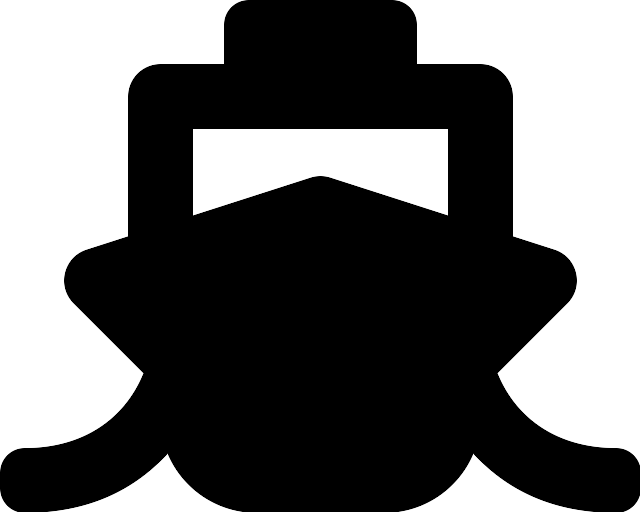}}
\newcommand{\faSheep}{\includegraphics[height=\fHeight]{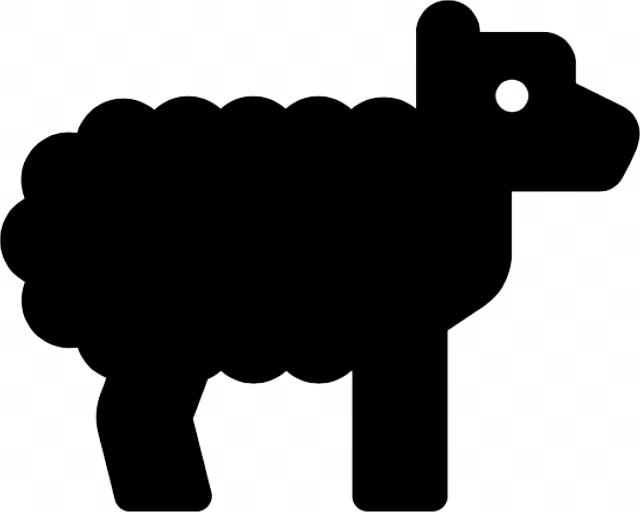}}

\newcommand{\faTrain}{\includegraphics[height=\fHeight]{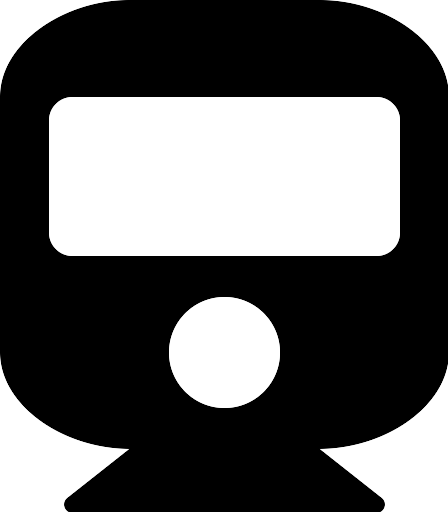}}
\newcommand{\faTulip}{\includegraphics[height=\fHeight]{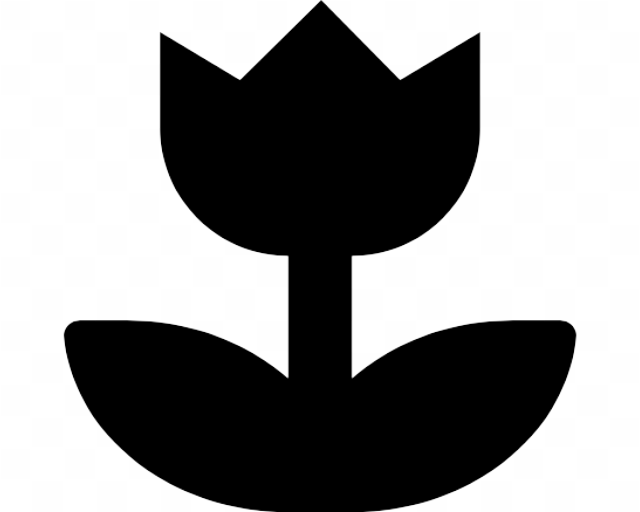}}
\newcommand{\faTv}{\includegraphics[height=\fHeight]{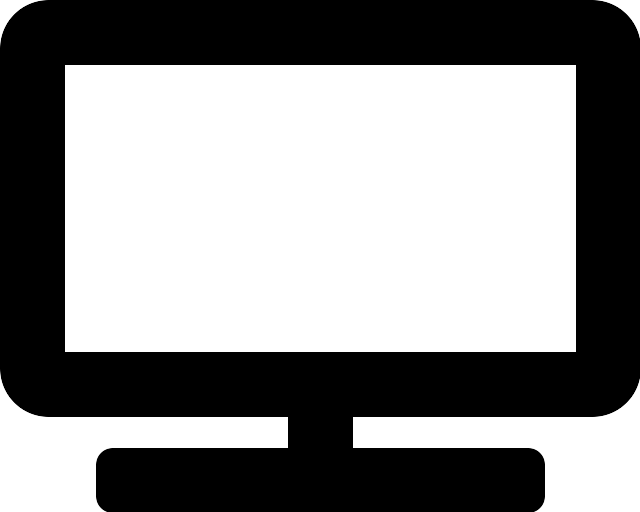}}
\newcommand{\faWalking}{\includegraphics[height=\fHeight]{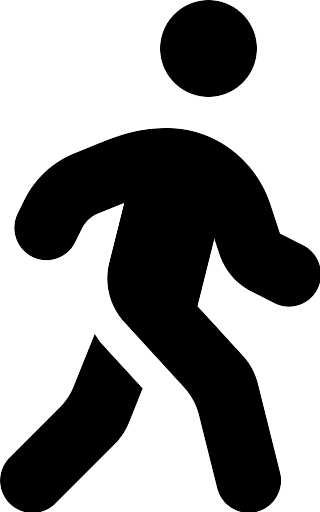}}
\newcommand{\faWineBottle}{\includegraphics[height=\fHeight]{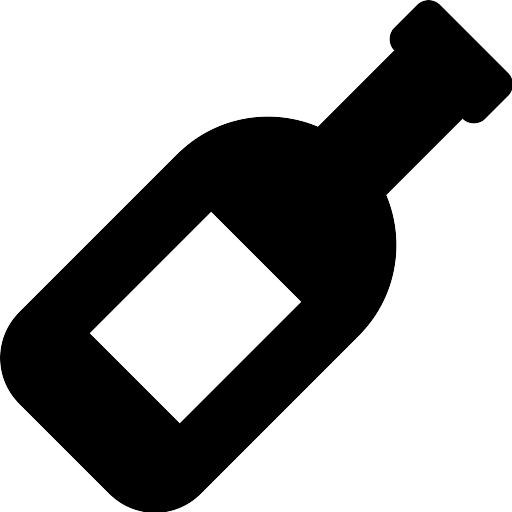}}

\newcommand{\faTiger}{\includegraphics[height=\fHeight]{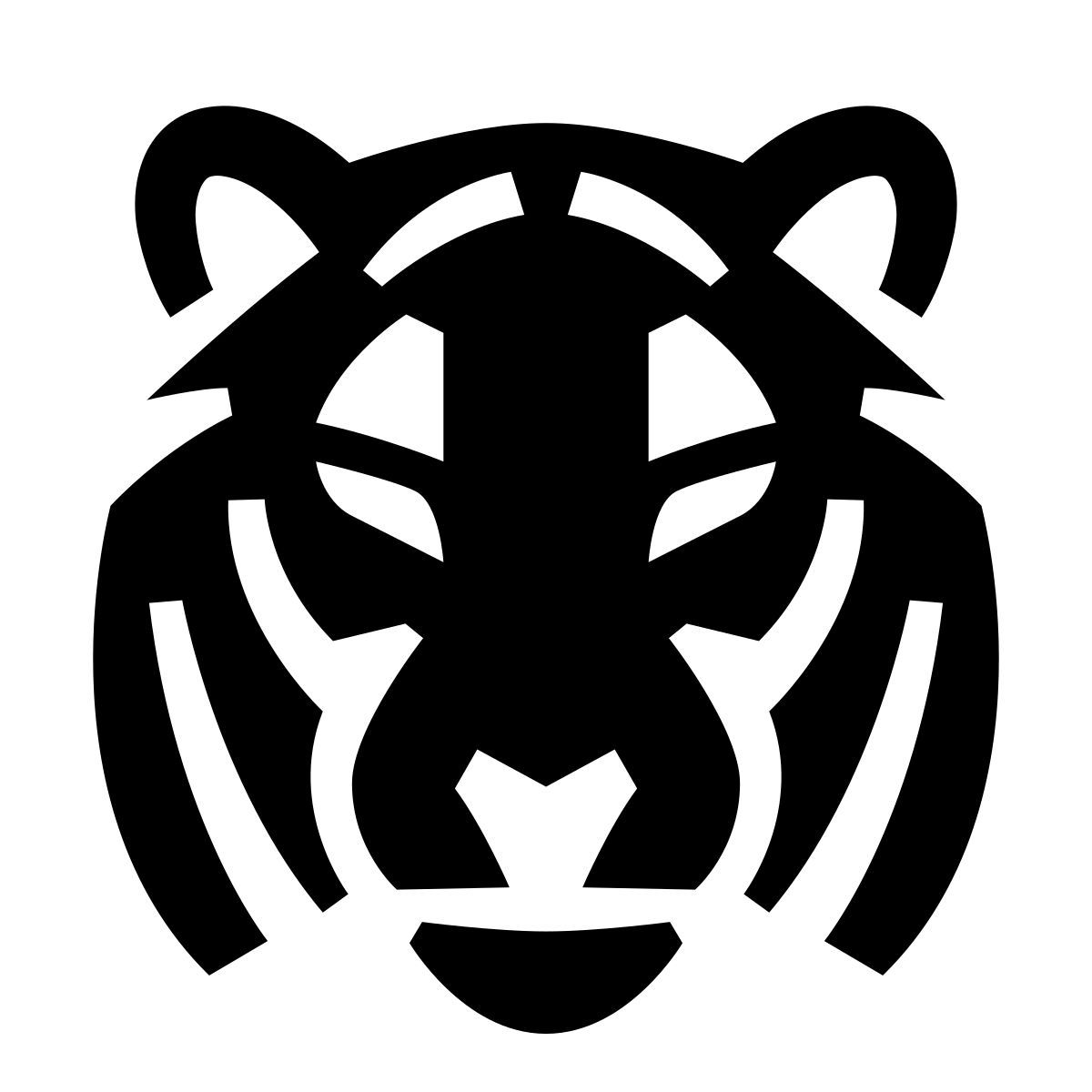}}
\newcommand{\faBear}{\includegraphics[height=\fHeight]{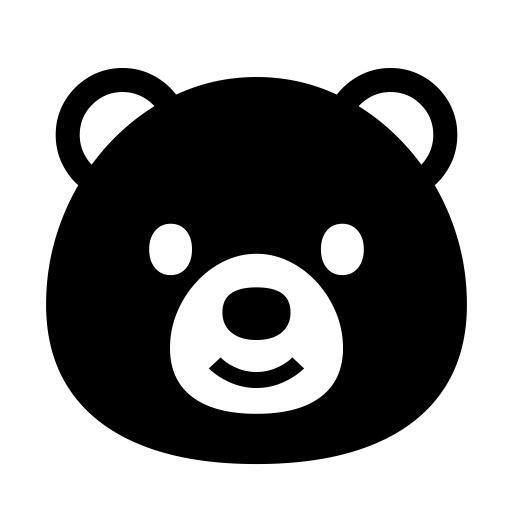}}
\newcommand{\faBuck}{\includegraphics[height=\fHeight]{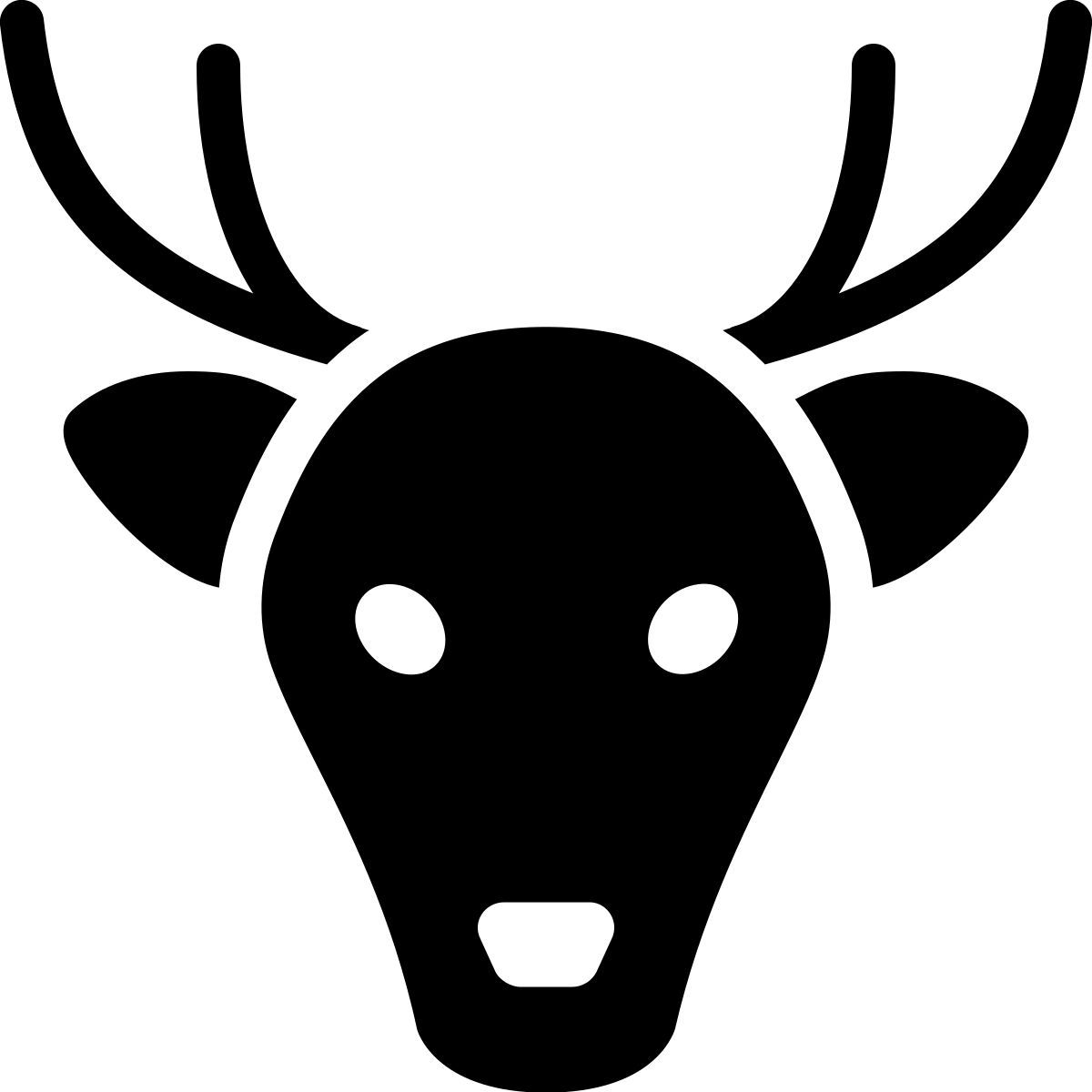}}
\newcommand{\faCattle}{\includegraphics[height=\fHeight]{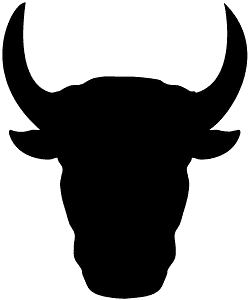}}
\newcommand{\faGiraffe}{\includegraphics[height=\fHeight]{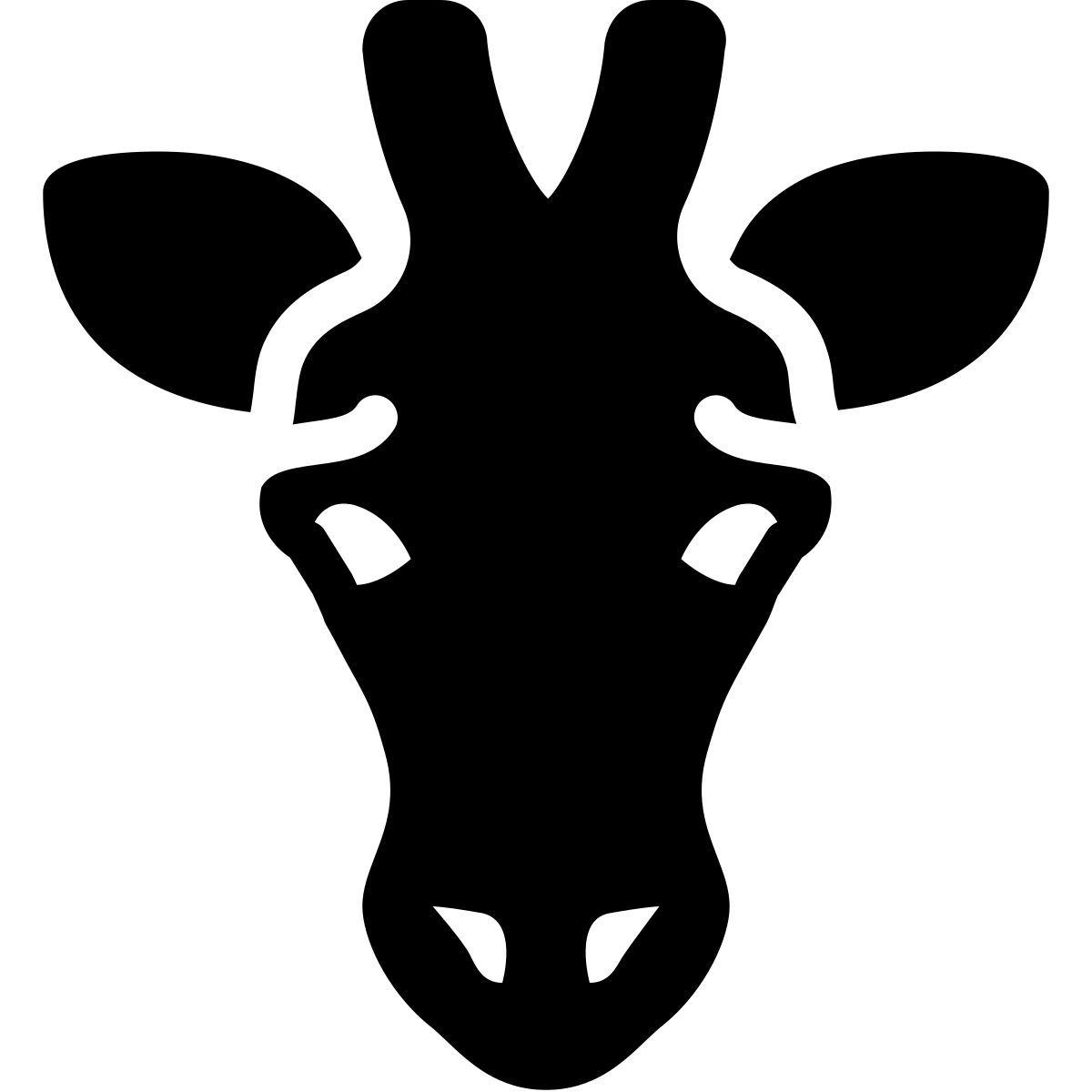}}
\newcommand{\faRhino}{\includegraphics[height=\fHeight]{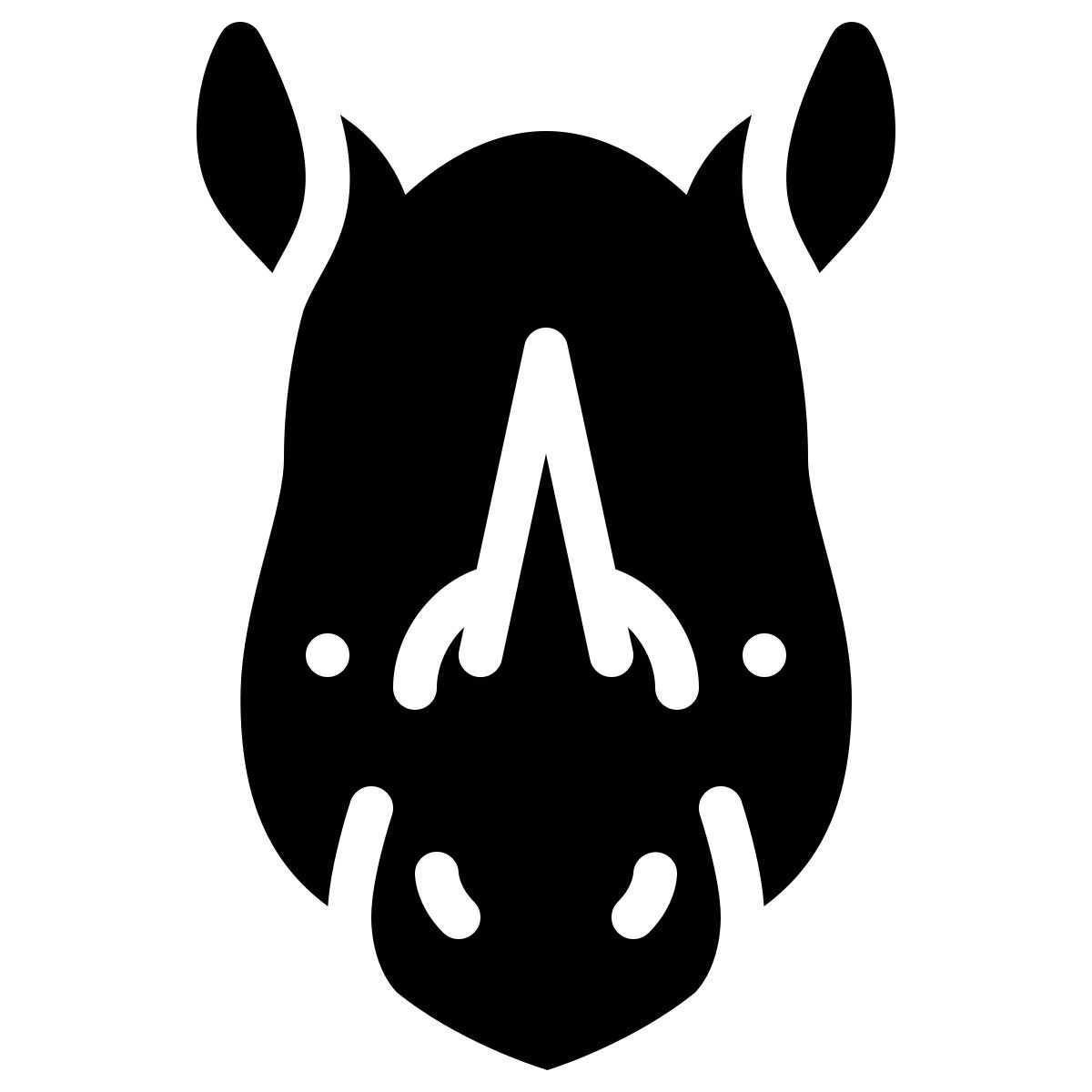}}

\newcolumntype{?}{!{\vrule width 1pt}}

\usepackage[accsupp]{axessibility}  %
\usepackage{hyperref}

\usepackage{orcidlink}

\begin{document}

\title{Unsupervised Pixel-Level Semantic Left-Right Understanding of In-the-Wild Images} 

\titlerunning{Pix2LR}

\author{Weikang Wang\inst{1,2} \and
        Tobias Wei{\ss}berg\inst{1,2} \and
        Florian Bernard\inst{1,2}}
\institute{University of Bonn, Germany \\
\and
Lamarr Institute, Germany}

\authorrunning{Weikang Wang et al.}

\maketitle

\begin{center}
    \captionsetup{type=figure}
    \setlength{\tabcolsep}{1.5pt}
\centering
\resizebox{\textwidth}{!}{\begin{tabular}{cccc} 
 Partiality & 
 Unbalanced left-right & Occlusion & 
 Complex boundaries \\
 \adjustbox{valign=m}{\includegraphics[height=0.2\textheight]{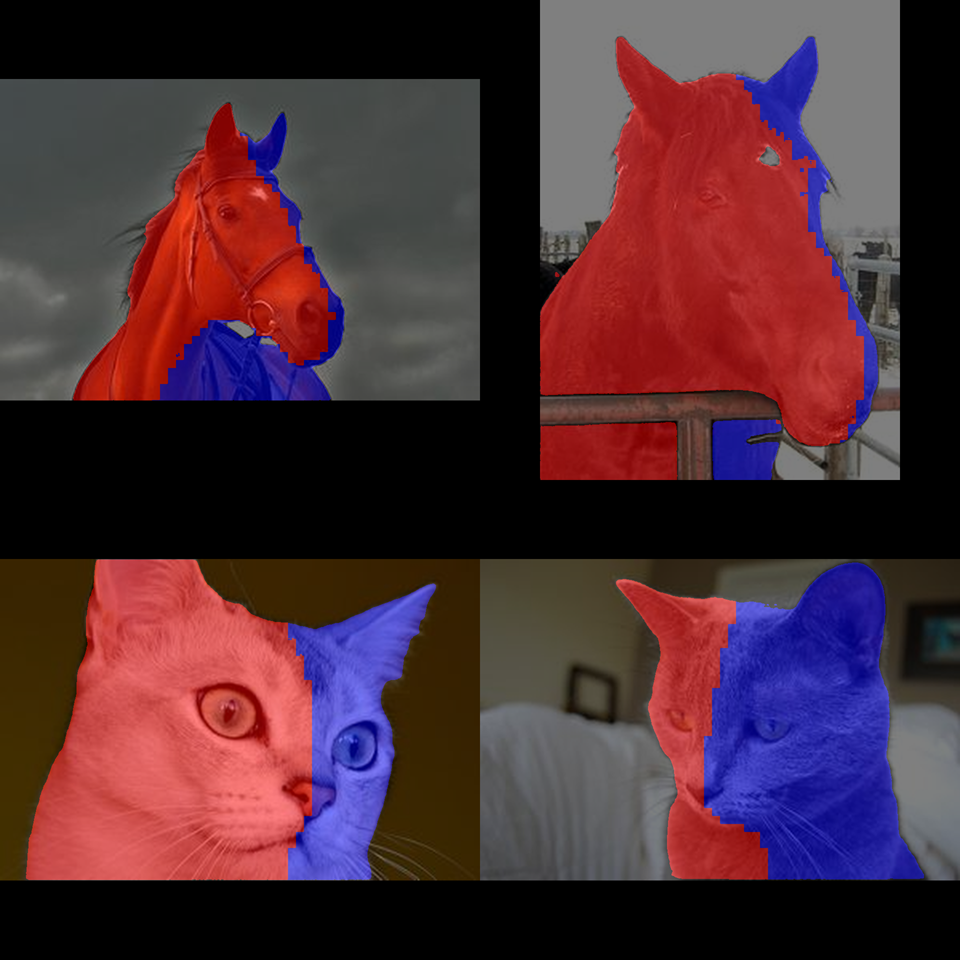}} &
\adjustbox{valign=m}{\includegraphics[height=0.2\textheight]{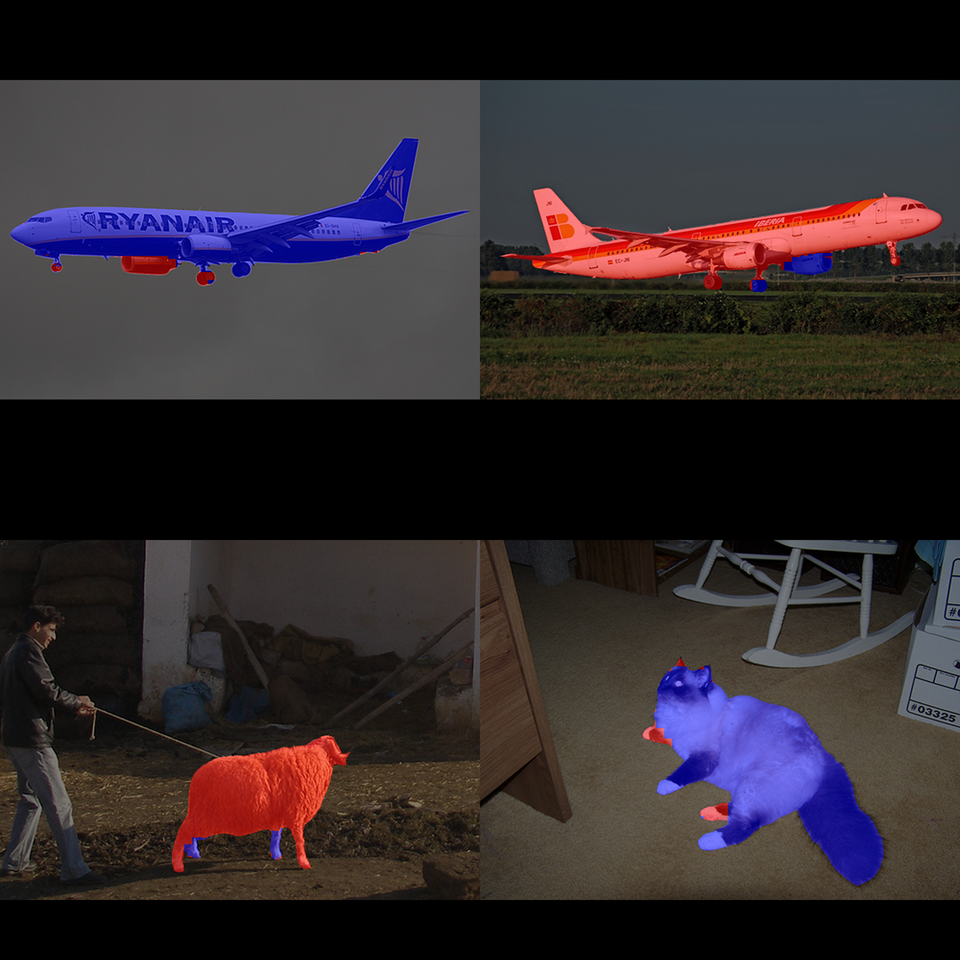}} &
\adjustbox{valign=m}{\includegraphics[height=0.2\textheight]{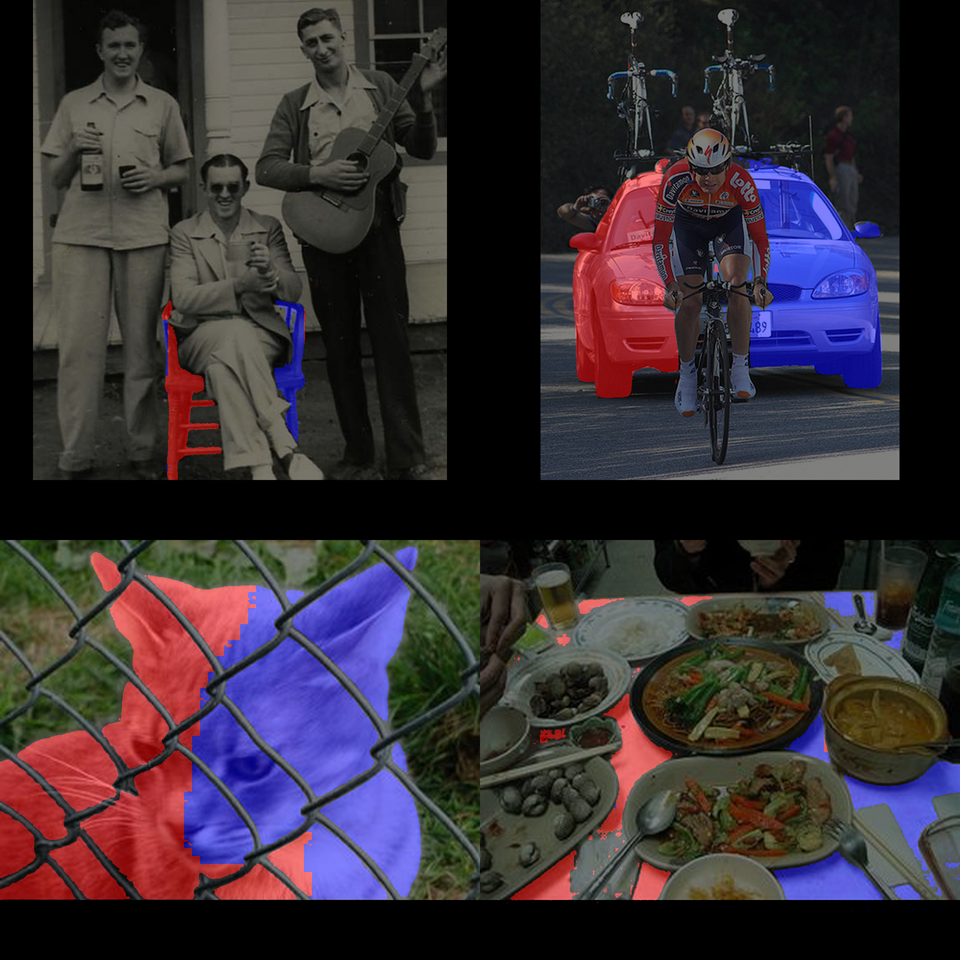}} & 
\adjustbox{valign=m}{\includegraphics[height=0.2\textheight]{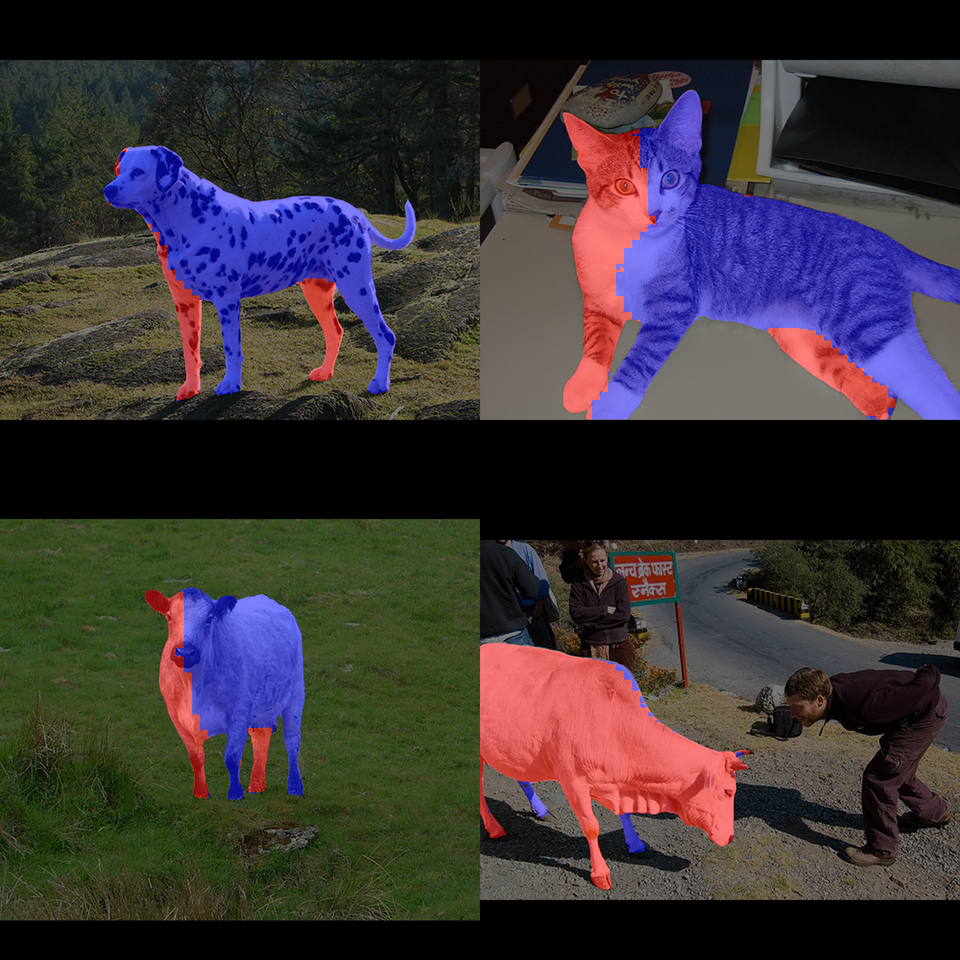}} \\[10ex]
Different scales & Unseen categories & Intra-class consistency & Inter-class consistency \\
\adjustbox{valign=m}{\includegraphics[height=0.2\textheight]{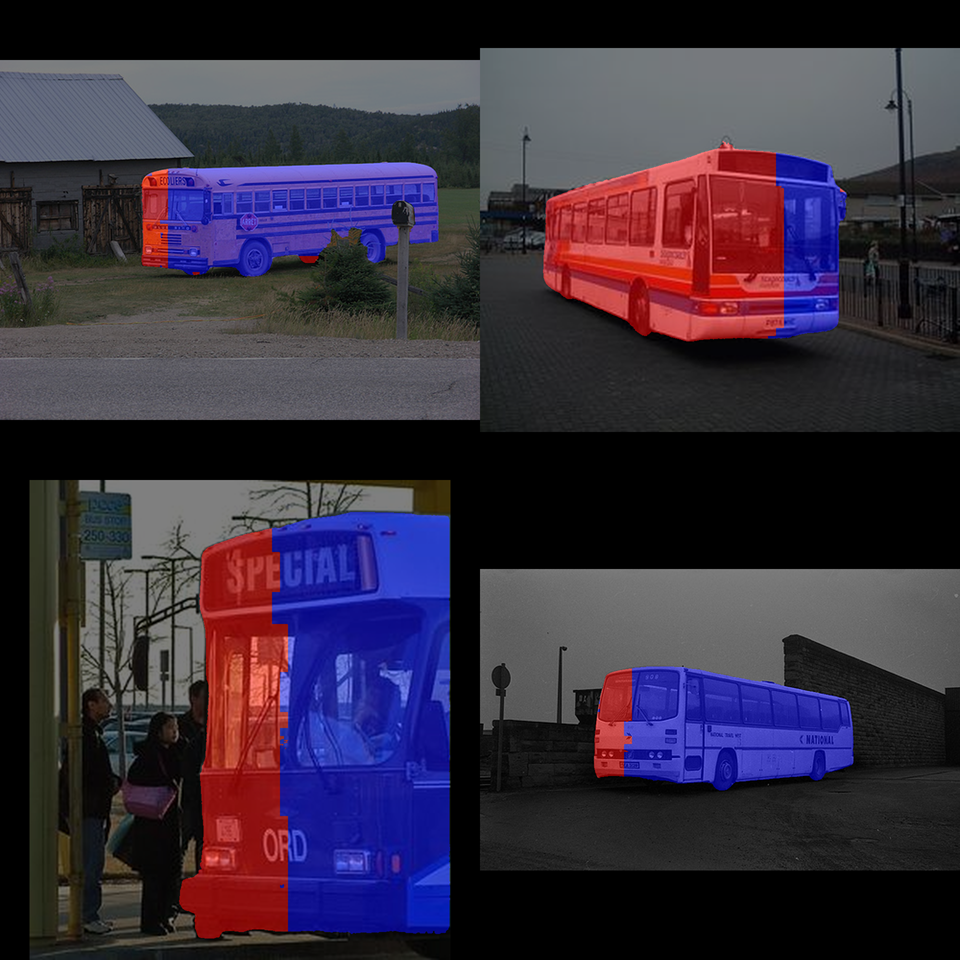}} &
\adjustbox{valign=m}{\includegraphics[height=0.2\textheight]{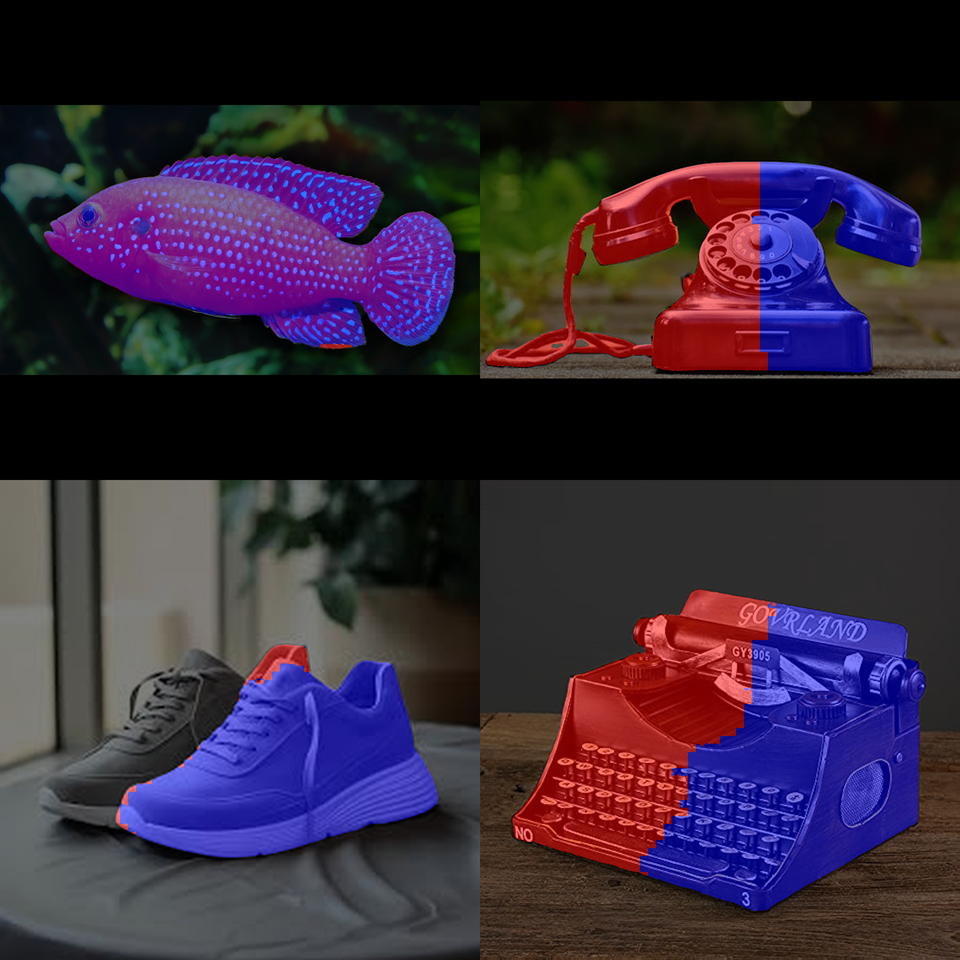}} &
\adjustbox{valign=m}{\includegraphics[height=0.2\textheight]{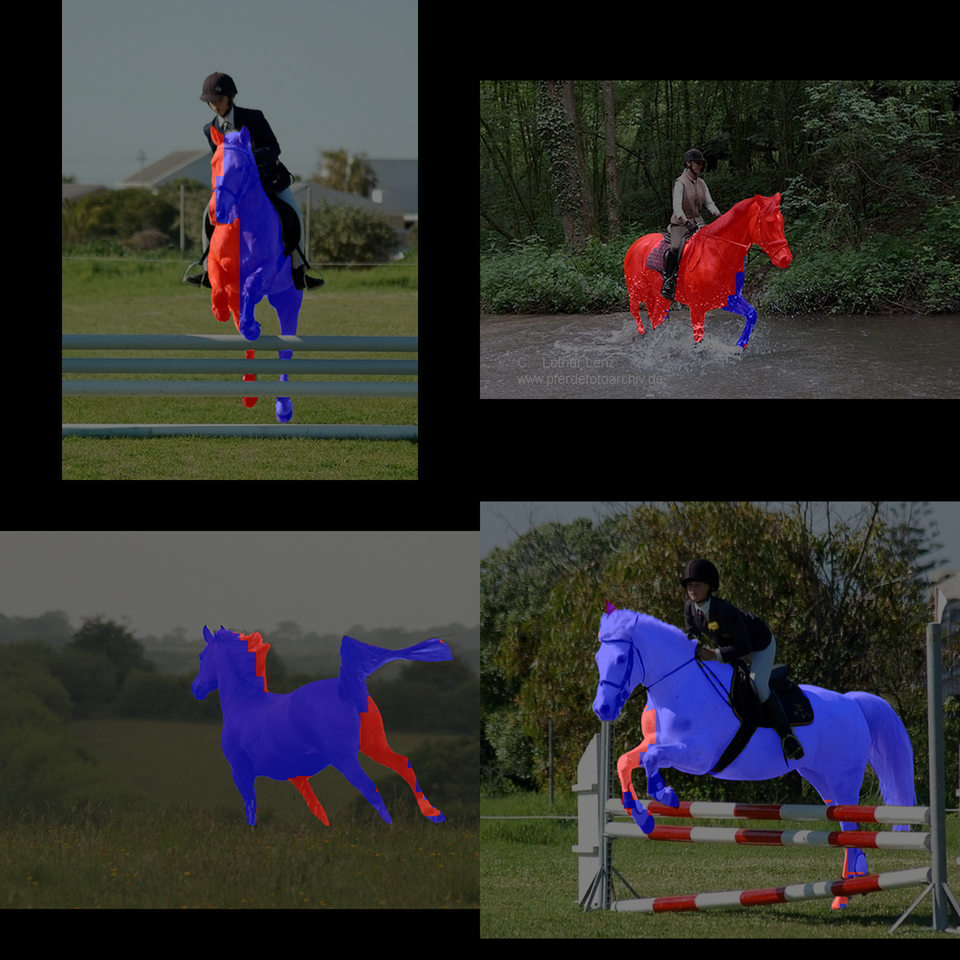}} &
\adjustbox{valign=m}{\includegraphics[height=0.2\textheight]{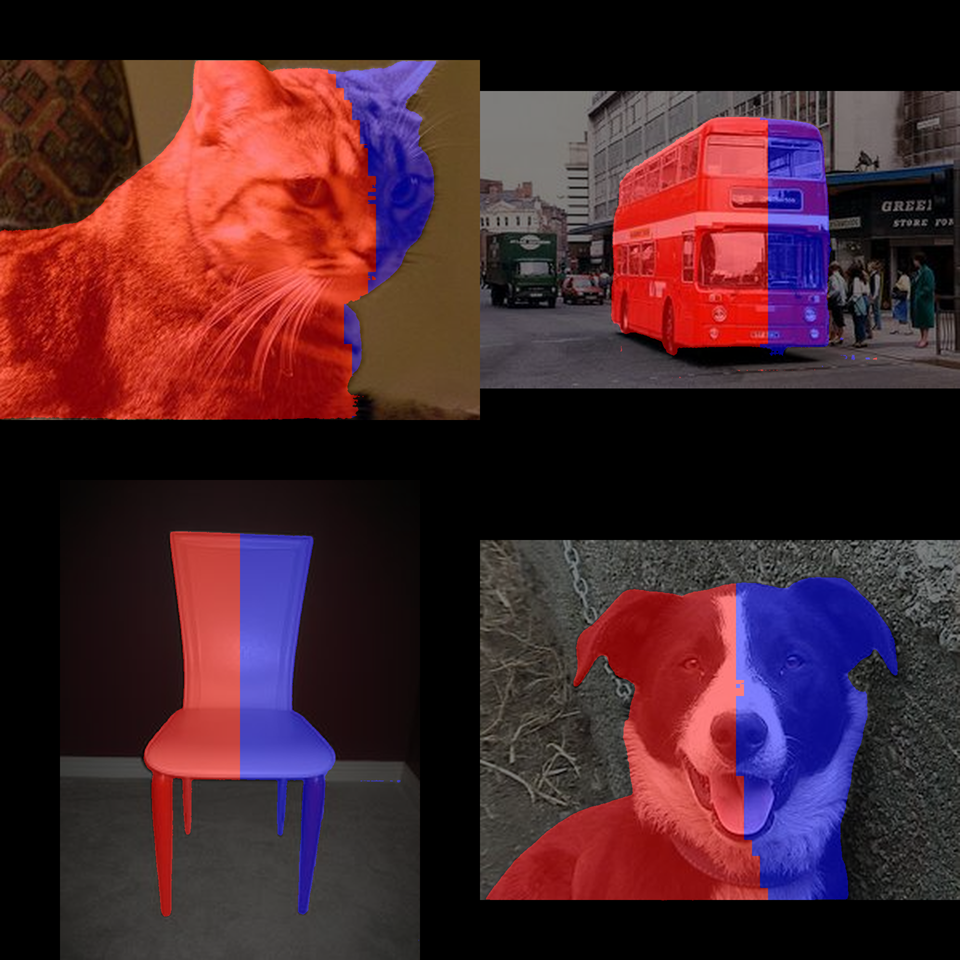}}
\end{tabular}}
    \caption{We propose the first unsupervised \textit{pixel-level} semantic left-right prediction framework for in-the-wild images, which is robust across diverse challenging settings.
}
    \label{fig:teaser}
\end{center}

\begin{abstract}
\label{abstract}

While various works address reflective symmetry understanding in 3D data and images, pixel-level semantic left-right prediction of in-the-wild images remains challenging, due to certain difficulties including the lack of 3D information, occlusion, object pose variation, partiality, \etc. In this work, we propose an unsupervised learning framework to tackle this challenge. Leveraging recent advances in vertex-wise semantic left-right understanding of 3D data, our unsupervised learning method jointly utilises 3D shape and image datasets to infer
pixel-wise semantic left-right predictions in single-view images.
In particular, we show that a medium-scale 3D shape dataset comprising mainly of human- and quadruped animal-like shapes, combined with diverse in-the-wild image data, are sufficient to achieve high-quality  semantic left-right prediction in images, even for entirely unseen 3D object categories, such as cars or trains. Overall, our approach achieves superior performance in dense pixel-wise semantic left-right predictions on both rendered and in-the-wild image datasets when compared to existing state-of-the-art methods.

\end{abstract}
\section{Introduction}
\label{sec:intro}

Reflective symmetry understanding, especially left-right symmetry understanding, has been a long standing topic that is studied in different areas of visual computing. The high relevance of left-right symmetry stems from the fact that it is a ubiquitous property observed in various object categories, including humans, animals and man-made objects such as cars, bicycles, or aeroplanes.

From a \textit{geometric} perspective, reflective symmetries (including left-right symmetry) are conventionally classified as extrinsic or intrinsic ones~\cite{liu2010computational, mitra2013symmetry}. 
For 3D data, such as point clouds and meshes, numerous methods capable of detecting extrinsic \cite{wang2024key, li2025symmetry, aguirre2025dataset, je2024robust} or intrinsic reflective symmetries~\cite{liu2012finding, mitra2013symmetry, ovsjanikov2008global, kim2010mobius, qiao2022learning, wang2025kh} have demonstrated high-quality results. 

In contrast, left-right understanding in 2D images is significantly more challenging. Left-right symmetry observed in images arises from the underlying 3D structure of the object they depict, and common difficulties including the lack of 3D information in single-view images, occlusion, pose variation and partiality, make geometric modelling of symmetries in 2D images ill-posed in many settings. 

Existing methods attempt to address this challenge, but they possess certain limitations. One category of methods focuses on general symmetry axis detection \cite{yu2025axis, seo2021learning, seo2022reflection, yang2025clipsym}, yet they are restricted to objects exhibiting (near) perfect extrinsic symmetry and struggle to handle cases like occlusion or partiality. Another line of works \cite{hartwig2025geco, zhang2024telling, wandel2025semalign3d} utilises semantic correspondence as a proxy to refine left-right aware image features; however, keypoints are required as supervision signals.

Recent works \cite{wang2025kh, weissberg2026symmetry} introduce a new \textit{semantic} perspective to tackle left-right understanding problem. Emerging studies \cite{zhang2024telling, cheng2024zero, el2024probing, mariotti2026jamais} show vision foundation models encode rich left-right semantics. Leveraging this, \cite{wang2025kh, weissberg2026symmetry} formalise left-right understanding of 3D shapes as a vertex-wise semantic prediction task. By utilising shape descriptors decorated with features from vision foundation models \cite{dutt2024diffusion}, these methods extract reliable predictions without requiring annotated data, effectively introducing a semantic paradigm to left-right understanding.

Inspired by this progress in vertex-wise left-right understanding of 3D shapes, 
we propose Pix2LR, a flexible and straightforward unsupervised framework for pixel-wise semantic left-right prediction, to \textbf{explicitly detect the semantic left and right parts} of objects within images. We summarise our main contributions as follows:

\begin{itemize}
    \item We propose the first unsupervised framework to predict dense pixel-level semantic left-right labels across a broad range of objects in in-the-wild images.
    \item To achieve this, we leverage a hybrid training that uses a medium-scale 3D shape dataset of limited object categories (human- and animal-like shapes), together with image datasets of a broader class of object categories.
    \item We design a per-vertex left-right prediction aggregation strategy to ensure the consistency for left-right predictions of shape and image data.
    \item We show the effectiveness of our dense pixel-level predictions for semantic left-right understanding, both on rendered and in-the-wild image datasets.
    \item Evaluation on unseen categories and image styles show the strong generalisation ability of our proposed framework.
\end{itemize}
\section{Related Works}
\label{sec:related_works}
We briefly review works that are closely related to reflective symmetry and left-right understanding.

\textbf{Left-Right understanding in 3D domain.} 
Traditional symmetry detection methods that geometrically model reflective symmetries as invariant to reflection (extrinsic symmetries) or isometries (intrinsic symmetries) can be directly leveraged for left-right understanding of 3D data. In the most straightforward setting, extrinsic symmetry detection methods have already achieved high-quality results. Mitra \etal \cite{mitra2006partial} propose an algorithm that matches simple local shape signatures in pairs and uses these matches to accumulate evidence for symmetries in an appropriate transformation space, followed by clustering and verification steps. Je \etal \cite{je2024robust} improve upon \cite{mitra2006partial} by utilizing langevin dynamics within an altered symmetry space to eliminate the verification step. E3Sym \cite{li2023e3sym} achieves symmetry detection by extracting E(3)-invariant features through a lightweight neural network, facilitating robust symmetry predictions.

As a more challenging task, intrinsic symmetry detection has garnered attention in recent decades. Ovsjanikov \etal \cite{ovsjanikov2008global} model intrinsic shape symmetries as extrinsic symmetries in eigenfunction signature space and detect symmetries by searching over all possible candidates. Nagar \etal \cite{nagar2018fast} improve this algorithm by sampling sparse intrinsic symmetric vertex pairs and leveraging the fact that the shortest geodesic path between two intrinsically symmetric points is also intrinsically symmetric. Liu \etal \cite{liu2012finding} extract intrinsic symmetries on genus-zero meshes by extracting closed curves from conformal maps of extremal points.

Recently, methods specifically designed for semantic left-right understanding of 3D data have been proposed. Using shape vertex descriptors enriched with semantic features from vision foundation models \cite{dutt2024diffusion}, $\chi$ \cite{wang2025kh} detects per-vertex left-right symmetry values across different shape categories using a combination of geometric losses. Wei\ss{}berg \etal \cite{weissberg2026symmetry} extend \cite{wang2025kh} by disentangling input vertex-based features into left-right agnostic and informative components, and improve the robustness of detected left-right symmetries.

\textbf{Left-Right understanding in 2D domain.} 
Due to the absence of 3D information in single-view 2D images, left-right understanding in 2D images is significantly more challenging compared to its 3D counterpart. While existing methods attempt to address this challenge, they suffer from various notable limitations. Several works \cite{seo2021learning, seo2022reflection, yu2025axis} target at general symmetry axis detection. Seo \etal \cite{seo2021learning} detect symmetries by employing polar pooling, self-similarity encoding, and angle-specific kernels, combined with a self-supervised data augmentation technique. Equisym \cite{seo2022reflection} uses an end-to-end framework for symmetry detection that leverages dihedrally-equivariant feature maps. Yu \etal \cite{yu2025axis} introduce a novel framework for axis-level extrinsic symmetry detection by representing symmetry axes as explicit geometric primitives, namely lines. While these symmetry axis prediction methods detect reasonable extrinsic symmetries and can be applied to left-right symmetry understanding, they exhibit poor performance on objects with occlusions, partialities, complex poses, and other challenging real-world conditions.

Another line of works \cite{zhang2024telling, hartwig2025geco, wandel2025semalign3d} in the semantic correspondence community aim to fine-tune features from vision foundation models to be more left-right aware, typically by using annotated sparse keypoint pairs as supervisory signals. Although the refined features perform well in semantic correspondence, they are not capable of performing pixel-wise left-right understanding effectively.

\textbf{Left-Right information from vision foundation models.} Vision foundation models, such as DINOv2 \cite{oquab2024dinov2}, DINOv3 \cite{simeoni2025dinov3}, CLIP \cite{radford2021learning}, and Stable Diffusion \cite{rombach2022high}, have demonstrated remarkable effectiveness across various image-based tasks.
As a ubiquitous property spanning different object categories, left-right information embedded within vision foundation models has been analysed in recent research.
El Banani \etal \cite{el2024probing} demonstrate that directly utilising features from vision foundation models for certain tasks (\eg, semantic correspondence) can sometimes lead to incorrect matching of similar structures (including left-right symmetric parts) while succeeding in other cases. This validates the existence of noisy left-right information within these features.
Cheng \etal \cite{cheng2024zero} optimise a spectral domain mapping derived from features of one vision foundation model to achieve consensus with features from another vision foundation model. Their results of correct left-right keypoint matching between images provide evidence for the existence of left-right information within vision foundation models.
Zhang \etal \cite{zhang2024telling} quantitatively validate the performance of different vision foundation models \cite{oquab2024dinov2, rombach2022high} for object left-right pose estimation by computing similarities between features of input images and manually designed pose templates.

As previously discussed, two recent works on 3D shape semantic left-right understanding \cite{wang2025kh, weissberg2026symmetry} have successfully extracted left-right information from per-vertex features enriched with vision foundation models. In this work, operating under the similar assumption that vision foundation models contain rich left-right information, we employ two widely-used vision foundation models: Stable Diffusion \cite{rombach2022high} (abbreviated as SD) and DINOv3 \cite{simeoni2025dinov3}.

To our knowledge, no existing method directly addresses pixel-wise left-right understanding of images semantically. Given the evidence that vision foundation models contain rich left-right semantics \cite{zhang2024telling, cheng2024zero, el2024probing}, and inspired by semantic left-right understanding of 3D data \cite{wang2025kh, weissberg2026symmetry}, this work aims to bridge this gap.
\section{Method}
\label{sec: method}

We first briefly recap the main ideas of $\chi$ \cite{wang2025kh}, a recent state-of-the-art unsupervised method for vertex-wise semantic left-right prediction of intrinsically symmetric 3D shapes. Then, we elaborate (i) the hybrid training data preparation of unannotated 3D shape and in-the-wild image datasets, (ii) the dense pixel-wise left-right predictor, (iii) the per-vertex left-right prediction aggregation process, and (iv) a combination of unsupervised losses, of our pixel-wise semantic left-right prediction framework, Pix2LR. Our notation is summarised in \cref{tab:symbols}.

Implementation details, including detailed network structure and training details, are provided in the supplementary material.

\begin{table}
\setlength{\tabcolsep}{5pt}
  \centering
  \caption{Summary of the notation used in this paper.}
  \label{tab:symbols}
  \resizebox{\textwidth}{!}{\begin{tabular}{@{}llll@{}}
    \toprule
    \textbf{Symbol} & \textbf{Description} & \textbf{Symbol} & \textbf{Description} \\
    \midrule
    $\mathcal{M}=(V,E)$ & Triangle mesh & $F_r \in \mathbb{R}^{N_r \times H \times W \times D}$ & Features of $I_r$ \\
    $V_{\mathcal{M}}$ & Vertices of $\mathcal{M}$ & $\bar{F}_r \in \mathbb{R}^{N_r \times H \times W \times D}$ & Flipped features of $\bar{I}_r$ \\
    $E_{\mathcal{M}}$ & Edges of $\mathcal{M}$ & $F_w \in \mathbb{R}^{N_w \times H \times W \times D}$ & Features of $I_w$ \\
    $I_r \in \mathbb{R}^{N_r \times H \times W \times 3}$ & Rendered images & $\bar{F}_w \in \mathbb{R}^{N_w \times H \times W \times D}$ & Flipped features of $\bar{I}_w$ \\
    $\bar{I}_r \in \mathbb{R}^{N_r \times H \times W \times 3}$ & Flipped rendered images & $\mathcal{F}_\mathcal{M}, \bar{\mathcal{F}}_\mathcal{M} \in \mathbb{R}^{\vert V_{\mathcal{M}} \vert \times D}$ & Per-vertex features of $\mathcal{M}$  \\
    $I_w \in \mathbb{R}^{N_w \times H \times W \times 3}$ & In-the-wild images & $S_w, \bar{S}_w \in [-1,1]^{N_w \times H \times W}$ & Per-pixel left-right values \\
    $\bar{I}_w \in \mathbb{R}^{N_w \times H \times W \times 3}$ & Flipped in-the-wild images & $\chi_\mathcal{M}, \bar{\chi}_\mathcal{M} \in [-1,1]^{\vert V_{\mathcal{M}} \vert}$ & Per-vertex left-right values \\
    $M_w \in \{0,1\}^{N_w \times H \times W}$ &  Object masks of $I_w$ &  $\chi_v, \bar{\chi}_v \in [-1,1]$ & Left-right value of vertex $v$ \\
    \bottomrule
  \end{tabular}
  }
\end{table}

\subsection{Background}
\label{sec: background}

First, $\chi$ \cite{wang2025kh} follows Diff3F \cite{dutt2024diffusion} to obtain vertex-wise shape features by aggregating image features from vision foundation model onto the 3D shape surface.  
Specifically, with an input triangle mesh $\mathcal{M}=(V_{\mathcal{M}},E_{\mathcal{M}})$, $N_r$ images are rendered from surrounding virtual cameras. And then the feature descriptor of each vertex is defined as the averaged image features of corresponding pixels across all $N_r$ views (vertex-to-pixel correspondences are known due to the rendering process), forming per-vertex shape features $\mathcal{F}_{\mathcal{M}} \in \mathbb{R}^{\lvert V_{\mathcal{M}} \rvert \times D}$.
Subsequently, $\chi$~\cite{wang2025kh} (i) flips each rendered image \textit{horizontally} to create a virtual image (with swapped roles of left and right), (ii) extracts image features of those flipped images, (iii) reverts the horizontal flip of the feature images, and then aggregates the image-based features to get per-vertex feature descriptors $\bar{\mathcal{F}}_{\mathcal{M}} \in \mathbb{R}^{\lvert V_{\mathcal{M}} \rvert \times D}$.
And finally, a lightweight network is utilised to learn a projection of the high-dimensional per-vertex features onto a one-dimensional chirality subspace (for both $\mathcal{F}_{\mathcal{M}}$ and $\bar{\mathcal{F}}_{\mathcal{M}}$), denoted as $\chi_{\mathcal{M}}, \bar{\chi}_{\mathcal{M}} \in [-1,1]^{\lvert V_{\mathcal{M}} \rvert}$, respectively. By carefully designed losses that enforce left-right consistency between $\chi_{\mathcal{M}}$ and $\bar{\chi}_{\mathcal{M}}$ and certain geometrical constraints, it's demonstrated that no supervision is necessary to perform reliable per-vertex semantic left-right predictions. We summarise geometric losses that are relevant to this work in the the following.
The \emph{dissimilarity loss}
    \begin{equation}
    \mathcal{L}_{\text{dis}} = -\frac{1}{\sqrt{\lvert V_{\mathcal{M}} \rvert}} \lVert \chi_{\mathcal{M}} - \bar{\chi}_{\mathcal{M}} \rVert_2
    \label{eq: dis}
\end{equation}
aims to enforce a large difference in the left-right predictions of intrinsically symmetric vertex pairs.
The \emph{total variation loss}
\begin{equation}
    \mathcal{L}_{\text{var}} = \frac{1}{\lvert E_{\mathcal{M}} \rvert} \sum_{(u,v) \in E_{\mathcal{M}}} \lVert \chi_u - \chi_v \rVert_1 + \lVert \bar{\chi}_u - \bar{\chi}_v \rVert_1
    \label{eq: var}
\end{equation}
promotes spatial smoothness.
And the \emph{fifty-fifty loss}
\begin{equation}
    \mathcal{L}_{\text{fif}} = \frac{1}{\vert V_{\mathcal{M}} \vert} \left(\frac{\vert \chi^{\top}_{\mathcal{M}} \mathbf{1}_{|V|} \vert}{\lVert \chi_{\mathcal{M}} \rVert_{\infty}} + \frac{\vert \bar{\chi}^{\top}_{\mathcal{M}} \mathbf{1}_{|V|} \vert}{\lVert \bar{\chi}_{\mathcal{M}} \rVert_{\infty}} \right)
    \label{eq: fif}
\end{equation}
favours predictions in which left and right parts have the same number of assigned vertices.

\subsection{Pixel-wise Semantic Left-Right Prediction}
\label{sec: pix2sym}

Now we introduce our Pix2LR framework.

\begin{figure*}[ht!]
    \centering
    \includegraphics[width=\textwidth]{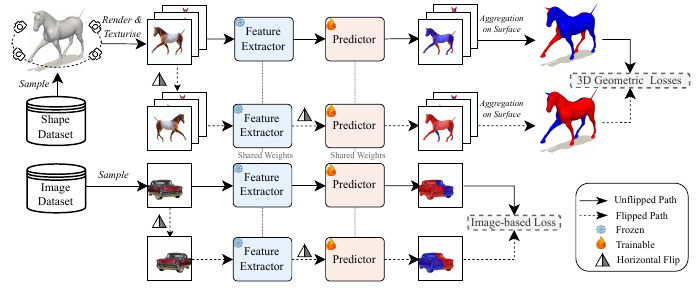}
    \caption{Overview of our pixel-wise semantic left-right prediction framework Pix2LR. We use a hybrid training dataset comprising of a medium-scale 3D shape dataset (mainly of human- and animal-like shapes), which we use to render images, together with an in-the-wild image dataset (of a broad class of object categories, \eg, including cars or aeroplanes). 
    Rendered and in-the-wild images are both flipped horizontally to create images with roles of left and right swapped. Then we compute feature descriptors using a pretrained (and fixed) feature extractor for all images. These feature descriptors are then fed into the Predictor to predict left-right per pixel. To train Pix2LR, we combine 3D geometric losses defined on the 3D shape,
    together with an image-based loss that enforces symmetric consistency between the original and the flipped image.
    }
    \label{fig:pipeline}
\end{figure*}

\textbf{Hybrid training data preparation.} We utilise a medium-scale 3D shape dataset together with in-the-wild image datasets for our training. Based on \cite{wang2025kh, weissberg2026symmetry, el2024probing, zhang2024telling, cheng2024zero}, semantic left-right information contained in vision foundation models are common and consistent for semantically left-right symmetric objects from different categories. And we hypothesize that the missing or incomplete left-right 3D information within in-the-wild images can be supplied by 3D shapes, even from other categories. To this end, we carefully design a hybrid training strategy. Below we first elaborate the hybrid training data preparation process. 

For each sampled 3D triangle mesh $\mathcal{M}=(V_{\mathcal{M}},E_{\mathcal{M}})$ from the shape dataset, we render $N_r$ images $I_r \in \mathbb{R}^{N_r \times H \times W \times 3}$ from surrounding views as in \cite{dutt2024diffusion, wang2025kh, weissberg2026symmetry}, and extract the features of those images using vision foundation models, denoted as $F_r \in \mathbb{R}^{N_r \times H \times W \times D}$. 

Analogously, we use the same vision foundation models to extract features for $N_w$ in-the-wild images $I_w \in \mathbb{R}^{N_w \times H \times W \times 3}$, sampled from the in-the-wild image dataset, denoted as $F_w \in \mathbb{R}^{N_w \times H \times W \times D}$.
Here, we assume that object masks $M_w \in \lbrace 0, 1 \rbrace^{N_w \times H \times W}$ are available (e.g.~obtained via some pre-trained mask detector such as SAM~\cite{kirillov2023segment}) for those in-the-wild images $I_w$.

\begin{figure}[!ht]
    \centering
    \captionsetup[sub]{labelformat=parens, labelfont=up, font=small,
        justification=raggedright, singlelinecheck=false}
    
    \settowidth{\figAwidth}{\includegraphics[height=0.133\textheight]{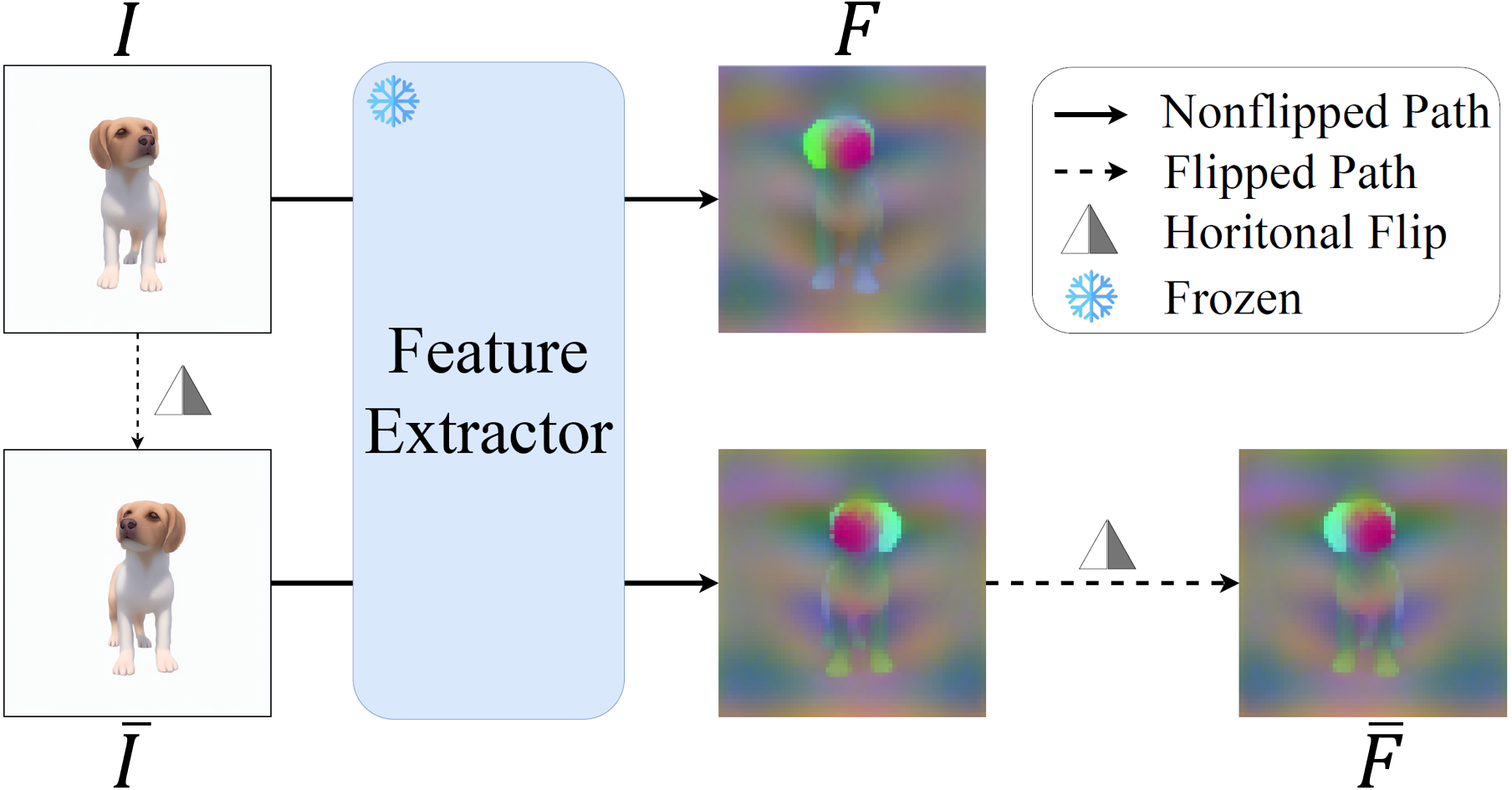}}
    \settowidth{\figBwidth}{\includegraphics[height=0.133\textheight]{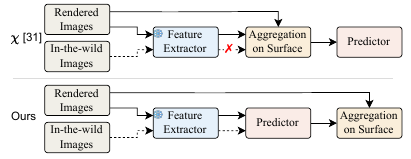}}
    
    \setlength{\tabcolsep}{4pt}
    \begin{tabular}{cc}
        \includegraphics[height=0.133\textheight]{images/flipping_big_font.png} &
        \includegraphics[height=0.133\textheight]{figures/aggregation.pdf} \\
    \end{tabular}
    
    \noindent
    \begin{minipage}[t]{\figAwidth}
        \subcaption{Feature extraction process.}
        \label{fig:flipping}
    \end{minipage}\hspace{2\tabcolsep}%
    \begin{minipage}[t]{\figBwidth}
        \subcaption{Comparison between $\chi$ and our method.}
        \label{fig:aggregation}
    \end{minipage}
    
    \caption{(a) Given an image $I$, we use a shared vision foundation model to extract features from both $I$ and its horizontally flipped counterpart $\bar{I}$. The feature obtained from $\bar{I}$ is then horizontally flipped back to align with $I$, yielding the final pair $(F, \bar{F})$. 
    (b) Solid and dashed lines denote the prediction paths of rendered and in-the-wild images, respectively. The top branch illustrates $\chi$ \cite{wang2025kh} and the bottom branch shows our method. Since aggregation on the surface requires pixel-to-vertex correspondences, which are only available for rendered images, the path from in-the-wild images to predictor is unavailable in $\chi$. Instead, we swap the order of left-right prediction and surface aggregation, so that our method keeps both paths unobstructed and produces consistent predictions for rendered and in-the-wild images.
    }
    \label{fig:flipping_and_aggregation}
\end{figure}

For any image either from rendered images set $I_r$ or in-the-wild images set $I_w$, we conduct the following steps: (i) horizontally flip the original image, (ii) then use the same vision foundation models as above to extract features of the flipped image, (iii) and finally revert the horizontal flip of the extracted features of the flipped image. \cref{fig:flipping_and_aggregation} (a) illustrates this process.

After gathering features of the original images $I_r,I_w$, and flipped features of flipped images $\bar{I}_r, \bar{I}_w$, we obtain each batch of training data: $(F_r, \bar{F}_r, F_w, \bar{F}_w)$.

In the above process, we adopt horizontal image flipping rather than vertical or arbitrary axis flipping for two reasons: (1) Any 2D-axis flip is equivalent to a horizontal flip (plus an image in-plane rotation) and thus results in the same chirality change for the geometry of underlying 3D object the image depicts (\ie, mirroring object in 3D space). (2) Our semantic features are extracted from vision foundation models whose training data overwhelmingly show upright objects (gravity bias). With that, these foundation models rarely see upside-down views (vertically flipped images), so the features of those views are less reliable. Previous works \cite{dutt2024diffusion, wang2025kh, weissberg2026symmetry, zhu2025densematcher} which leverage or extract features from vision foundation models, also incorporate this bias of camera positions. More explanations are in the supplementary materials.

\textbf{Pixel-wise left-right predictor $P_{\Phi}$.} We adopt a light-weight network as our pixel-wise left-right predictor $P_{\Phi}$, which takes image feature maps as input and outputs dense left-right predictions.
Specifically, For the input features $F \in \mathbb{R}^{H \times W \times D}$, which can either be from the rendered images or in-the-wild images, 
the dense left-right predictions $S$ is computed as $S = P_{\Phi}(F) \in [-1,1]^{H^{\prime} \times W^{\prime}}$,
where $H^{\prime}/H = W^{\prime}/W$ represents possible scaling. Details about predictor structure are in the supplementary materials.

\textbf{Per-vertex left-right prediction aggregation.} We feed features of $N_r$ rendered images, $F_r, \bar{F}_r$, into the predictor $P_{\Phi}$ to get pixel-wise left-right predictions $S_r, \bar{S}_r \in [-1,1]^{N_r \times H^{\prime} \times W^{\prime}}$.
Then, since vertex-to-pixel correspondence of each rendered image can be computed from virtual camera intrinsic and extrinsics, we decorate each vertex $v \in V_{\mathcal{M}}$ with the left-right value $\chi_v$ by averaging left-right predictions of its corresponding pixels from all $N_r$ views, \ie,
\begin{equation}
    \chi_v = \frac{1}{\sum_{n=1}^{N_r} \vert \mathcal{I}_n^v \vert} \sum_{n=1}^{N_r} \sum_{(x,y) \in \mathcal{I}_n^v} S_r[n,x,y],
    \label{eq: aggregation}
\end{equation}
where $\mathcal{I}_n^v$ denote the set of corresponding pixels of rendered image $n \in {1,2,\cdots, N_r}$ for vertex $v$. We stack the per-vertex left-right value $\chi_v \in [-1,1]$ of all vertices to form $\chi_{\mathcal{M}} \in [-1,1]^{\vert V_{\mathcal{M}} \vert}$. Similarly, $\bar{\chi}_{\mathcal{M}} \in [-1,1]^{\vert V_{\mathcal{M}} \vert}$ is computed using $\bar{S}_r$.

Note that, similar to $\chi$ \cite{wang2025kh}, we also obtain the per-vertex left-right values $\chi_{\mathcal{M}}, \bar{\chi}_{\mathcal{M}}$, but we do so in a different way by swapping the order of left-right prediction and aggregation. Right sub-figure in \cref{fig:flipping_and_aggregation} (b) explains the difference in details. By incorporating this new design, we enable the framework to perform semantic left-right understanding of different data modes (images and 3D triangle meshes) consistently, thus make joint training of hybrid data be feasible.

\textbf{Unsupervised losses.} To train our model, we use a combination of 3D geometric losses (imposed on the 3D shapes) and an image-based loss (imposed on the in-the-wild images). Specifically, we consider
\begin{equation}
    \mathcal{L} = \lambda_1 \mathcal{L}_{\text{dis}} + \lambda_2 \mathcal{L}_{\text{var}} + \lambda_3 \mathcal{L}_{\text{fif}} + \lambda_4 \mathcal{L}_{\text{img}},
    \label{eq: losses}
\end{equation}
where $\mathcal{L}_{\text{dis}}$, $\mathcal{L}_{\text{var}}$ and $\mathcal{L}_{\text{fif}}$ are as defined in \cref{eq: dis}, \cref{eq: var} and \cref{eq: fif}, computed using $\chi_{\mathcal{M}}$ and $\bar{\chi}_{\mathcal{M}}$ from above. $\mathcal{L}_{\text{img}}$ is the image-based loss:
\begin{equation}
    \mathcal{L}_{\text{img}} = -\frac{1}{N_w} \lVert (S_w - \bar{S}_w) {\cdot} M_w \rVert_F,
\end{equation}
where $S_w, \bar{S}_w \in [-1,1]^{N_w \times H^{\prime} \times W^{\prime}}$ are the left-right predictions got from the predictor $P_{\Phi}$ with $F_w, \bar{F}_w$ as input.
\section{Experiments}
\label{sec: experiments}

In this section we experimentally analyse on various image datasets the
quality of pixel-wise left-right predictions from our proposed framework Pix2LR. 

\textbf{Shape datasets.} We use the recently proposed medium-scale shape dataset \textsc{BeCoS} \cite{ehm2025beyond}.
\textsc{BeCoS} \cite{ehm2025beyond} includes different human-like and quadruped animals shapes of various species with a train/test/val split of size 1975/284/274. It also provides consistent per-vertex ground truth left-right annotations that can be used for evaluation. In our ablation study, we also consider a popular human shape dataset \textsc{FAUST} \cite{bogo2014faust}, which contains 100 human shapes with train/test/val split size of 80/10/10. Since \textsc{FAUST} \cite{bogo2014faust} does not provide left-right annotations, we use respective annotations offered by \textsc{BeCoS} \cite{ehm2025beyond} framework. 

\textbf{Image datasets.} We use two common multi-category in-the-wild image datasets \textsc{PF-Pascal} \cite{ham2016} and \textsc{SPair-71k} \cite{min2019spair}.
\textsc{PF-Pascal} \cite{ham2016} contains images from 20 categories with 1001/506/521 train/test/val, with similar viewpoints and instance poses within each category. \textsc{SPair-71k} \cite{min2019spair} contains images with large intra-class pose variation, with train/test/val as 997/481/322 and 18 categories. 

In order to fully evaluate the performance of our proposed framework, we use both in-the-wild images from \textsc{PF-Pascal} \cite{ham2016} and \textsc{SPair-71k} \cite{min2019spair} test sets, as well as images rendered from \textsc{BeCoS} \cite{ehm2025beyond} test shapes. 

\begin{figure}[!ht]
    \setlength{\tabcolsep}{2.5pt}
\centering
    \begin{tabular}{c?c}
         
         \adjustbox{valign=m}{\includegraphics[height=0.125\textheight]{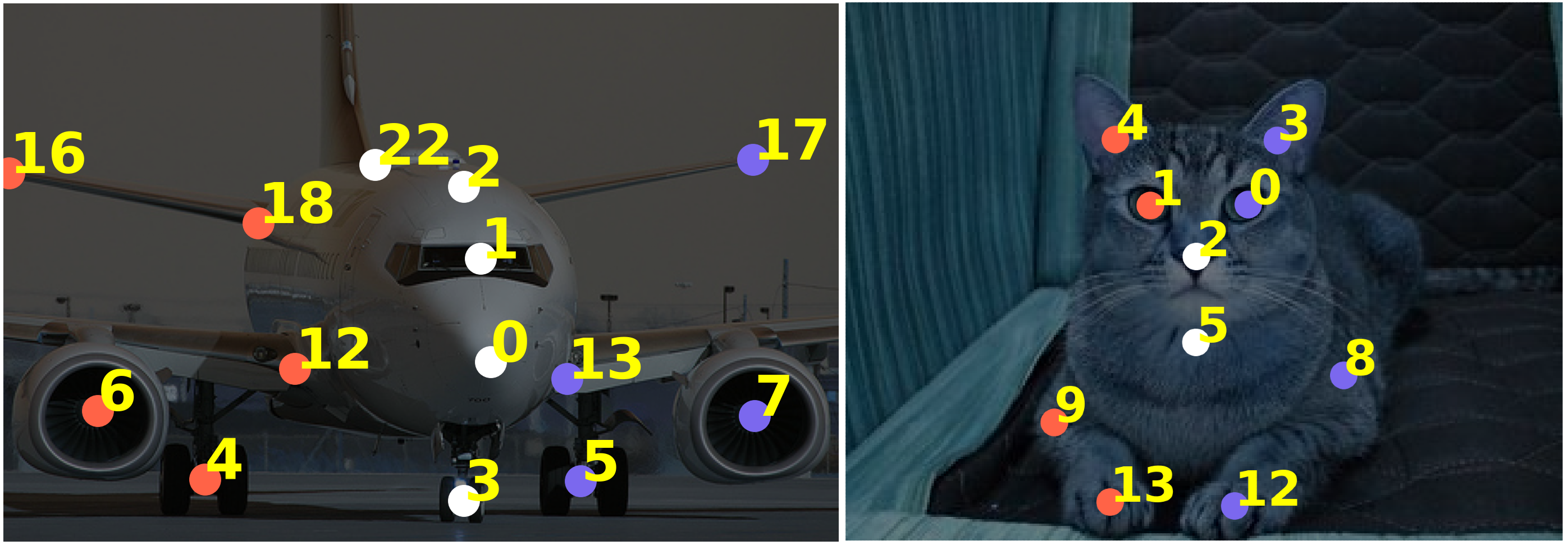}} &
         \adjustbox{valign=m}{\includegraphics[height=0.125\textheight]{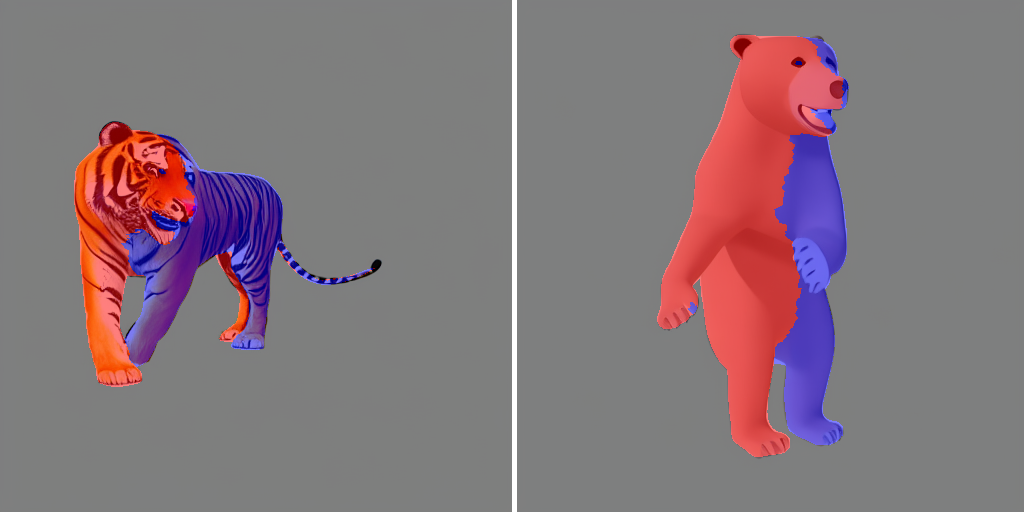}} \\ 

    \end{tabular}
    
    \caption{\textit{Left}: Examples of \textit{sparse} keypoint-wise left-right annotations of in-the-wild images from \textsc{SPair-71k} \cite{min2019spair} and \textsc{Pf-Pascal} \cite{ham2016} datasets. Blue and red denote points within left and right parts, and white denotes the points on left-right boundary. \textit{Right}: Examples of \textit{dense} pixel-wise left-right annotations in rendered images from \textsc{BeCoS} \cite{ehm2025beyond} dataset.
    Blue and Red denote semantic left and right parts.}
    \label{fig:new_annotation}
\end{figure}

For rendered images of \textsc{BeCoS} \cite{ehm2025beyond} test shapes, pixel-wise left-right annotations can be obtained from vertex-wise left-right annotations of shapes, as shown in the right sub-figure of \cref{fig:new_annotation}. For in-the-wild images from either \textsc{PF-Pascal} \cite{ham2016} or \textsc{SPair-71k} \cite{min2019spair}, left-right annotations are not available. By leveraging the consistent keypoint annotations within each category of these two datasets, we add new keypoint-wise left-right annotations to them. Specifically, for each category, we manually classify the keypoints into left side, right side and on the left-right boundary of each object (the left and right side is defined in the object view), and labelling them as $1$, $-1$ and $0$ respectively. Left sub-figure of \cref{fig:new_annotation} gives examples of these newly introduced keypoint-wise left-right annotations. More details about the annotation are in the supplementary materials.

\textbf{Metrics.} For either dense pixel-wise left-right annotations on rendered images from \textsc{BeCoS} \cite{ehm2025beyond}, or sparse keypoint-wise left-right annotations on in-the-wild images from \textsc{PF-Pascal} \cite{ham2016} and \textsc{SPair-71k} \cite{min2019spair}, we use the same accuracy metric as defined below. Specifically, for experiments over $N$ images:
\begin{equation}
    acc = \max(\frac{1}{N}{\sum_{i=1}^{N}}\frac{\mathbbm{1}(y_i {\cdot} y_i^{\text{gt}} > 0)}{ \lVert y_i {\cdot} y_i^{\text{gt}} \rVert_2^2}, \frac{1}{N}{\sum_{i=1}^{N}}\frac{\mathbbm{1}(y_i {\cdot} y_i^{\text{gt}} < 0)}{\lVert y_i {\cdot} y_i^{\text{gt}} \rVert_2^2}),
    \label{eq: acc_s}
\end{equation} where $y_i \in \{-1,1 \}^{M_i}$ and $y_i^{\text{gt}} \in \{-1,0,1 \}^{M_i}$ denote the predicted and ground-truth left-right values of all keypoints or all foreground pixels of image $i$, respectively. Note that the size $M_i$ depend on individual image $i$ . The final accuracy $acc$ is the maximum between two values above since the roles with left-right swapped are still valid as long as they are consistent within this category.

\textbf{Baselines.} We compare the performance of our framework with several state-of-the-art left-right aware feature descriptors: SD+DINOv2~\cite{zhang2023tale} and its variant SD+DINOv3, GeoAware \cite{zhang2024telling}, as well as a recent vertex-wise left-right predictor of 3D shapes $\chi$ \cite{wang2025kh}. Since $\chi$ \cite{wang2025kh} is originally designed to predict per-vertex left-right values independently, we adopt a pretrained $\chi$ \cite{wang2025kh} on \textsc{BeCoS} (using the same vision foundation models as ours for fair comparison), denoted as $\chi$\textsubscript{\scalebox{0.8}{BeCoS}}, to perform per-pixel left-right predictions with per-pixel features.

For each feature descriptor, different steps are conducted for in-the-wild images with sparse keypoint-wise left-right annotations, and for rendered images with dense pixel-wise left-right annotations, to obtain predictions. For in-the-wild images, the steps are as follows. For each category, (1). First, for each keypoint, compute its descriptor by averaging its corresponding features from all training images. Denote them as template keypoints.
(2). Then, apply a 2-centre clustering of all template keypoints without the ones labelled as $0$ (keypoints on the left-right boundary) using their descriptors, and label them with clustered labels as being -1 or 1 (keypoints on the left-right boundary labelled as 0). (3). Finally, for each keypoint within test images, assign it with the label of its closest template keypoint, in the sense of cosine similarity between its feature and template keypoint descriptors. For rendered images, the steps are as follows. As in \cite{dutt2024diffusion, wang2025kh, weissberg2026symmetry}, we decorate each shape vertex with the averaged image features of corresponding pixels from all $N_r$ rendered images. Then for each pixel within a test image, assign it with the left-right label of its closest vertex, in the sense of cosine similarity between features.

\subsection{Pixel-level Left-Right Prediction on Real Images}

For images within each category from \textsc{PF-Pascal} \cite{ham2016} or \textsc{SPair-71k} \cite{min2019spair}, we use the sparse keypoint-wise annotations defined as above, to compute the accuracy $acc$ over all test images within this category using \cref{eq: acc_s}, with $N$ being the number of images of this category, $K_c$ being the number of keypoints of this category, and $y_i \in \{-1,1 \}^{K_c}, y_i^{\text{gt}} \in \{-1,0,1 \}^{K_c}$ being the prediction and ground truth of all keypoints of image $i$, respectively.

\begin{table*}[tbh]
 \setlength{\tabcolsep}{0.5pt}
 \centering
 \caption{Left-Right predictions on \textsc{SPair-71k} \cite{min2019spair}. 
 Our method significantly outperforms its competitors for most object categories. For categories in which the competitive baseline GeoAware \cite{zhang2024telling} performs better, our method still gives comparable numbers. And for other categories and on average, our method outperforms GeoAware \cite{zhang2024telling} with a large margin. 
 We observe that training ours on both image dataset only leads to slightly better results on average compared to using only \textsc{SPair-71k} \cite{min2019spair} as image dataset.
 }
  \label{tab:spair-sparse}
 \footnotesize
\resizebox{\textwidth}{!}{\begin{tabular}{@{}l|cccccccccccccccccccc@{}}
    \toprule
    \textbf{Method} & \faPlane & \faBicycle & \faCrow & \faShip & \faWineBottle & \faBus & \faCar & \faCat & \faChair & \faCow & \faDog & \faHorse & \faMotorcycle & \faWalking & \faTulip & \faSheep & \faTrain & \faTv & $\stackrel{\textbf{AVG}}{acc}$ & $\hat{acc}$ \\ 
    \midrule
    SD+DINO~\cite{zhang2023tale} & 56.6 & 51.2 & 50.6 & 57.2 & 50.4 & 51.5 & 50.5 & 51.8 & 52.5 & 51.3 & 51.3 & 50.0 & 55.7 & 51.0 & 52.2 & 50.5 & 50.2 & 50.3 & 51.9 & 50.0 \\
    SD+DINOv3 & 55.2 & 51.2 & 50.6 & 57.2 & 50.9 & 50.8 & 50.8 & 51.4 & 52.9 & 51.8 & 50.1 & 50.0 & 56.7 & 51.0 & 96.9 & 50.5 & 50.5 & 96.1 & 57.2 & 55.8 \\
    GeoAware~\cite{zhang2024telling} & \textbf{96.5} & 53.1 & 50.6 & \textbf{57.9} & \textbf{90.4} & 51.8 & 50.5 & 51.4 & \textbf{89.7} & \textbf{95.6} & 51.3 & 51.0 & 70.3 & \textbf{94.4} & \textbf{100} & 50.5 & 50.1 & \textbf{100} & 70.1 & 64.0  \\
    $\chi$\textsubscript{\scalebox{0.8}{BeCoS}}~\cite{wang2025kh} & 72.2 & \textbf{69.4} & 79.8 & 55.5 & 73.4 & 72.0 & 61.9 & \textbf{84.5} & 73.0 & 80.5 & 79.3 & 73.9 & 53.0 & 82.6 & 75.5 & 74.8 & 79.1 & 84.1 & 73.6 & 73.0 \\
    Ours\textsubscript{\scalebox{0.8}{BeCoS-SPair71k}} & 76.9 & 67.4 & 81.0 & 52.0 & 89.7 & 93.4 & 89.6 & 83.2 & 86.7 & 85.0 & \textbf{81.1} & 76.2 & 71.0 & 89.2 & 98.2 & 76.0 & 90.4 & 99.0 & 82.7 & 82.7 \\
   Ours\textsubscript{\scalebox{0.8}{BeCoS-Both}} & 83.9 & 68.1 & \textbf{84.8} & 51.5 & 89.2 & \textbf{93.6} & 91.3 & 83.9 & 86.9 & 86.4 & 80.3 & \textbf{80.8} & \textbf{73.0} & \textbf{93.1} & 92.5 & \textbf{76.7} & \textbf{92.3} & 98.2 & \textbf{83.8} & \textbf{83.8} \\
    \bottomrule
  \end{tabular}}
\end{table*}
\begin{table*}[tbh]
 \setlength{\tabcolsep}{0.5pt}
 \centering
 \caption{Left-Right predictions on \textsc{PF-Pascal} \cite{ham2016}. 
 Our method significantly outperforms its competitors for most object categories. We observe that training ours on the \textsc{PF-Pascal} \cite{ham2016} image dataset only leads to slightly better results on average compared to using both \textsc{SPair-71k} \cite{min2019spair} and \textsc{PF-Pascal} \cite{ham2016} as image dataset.}
 \label{tab:pfpascal-sparse}
  \footnotesize
\resizebox{\textwidth}{!}{\begin{tabular}{@{}l|cccccccccccccccccccccc@{}}
    \toprule
    \textbf{Method} & \faPlane & \faBicycle & \faCrow & \faShip & \faWineBottle & \faBus & \faCar & \faCat & \faChair & \faCow & \faDinningTable & \faDog & \faHorse & \faMotorcycle & \faWalking & \faTulip & \faSheep & \faCouch & \faTrain & \faTv & $\stackrel{\textbf{AVG}}{acc}$ & $\hat{acc}$\\
    \midrule
    SD+DINO~\cite{zhang2023tale} & 55.4 & 63.0 & 60.4 & 50.0 & 50.0 & 50.3 & 53.5 & 52.0 & 51.3 & 50.0 & 50.8 & 50.0 & 50.9 & 63.8 & 50.1 & 50.0 & 50.0 & 53.5 & 52.5 & 51.5 & 54.1 & 51.9  \\
    SD+DINOv3 & 55.4 & 61.0 & 60.4 & 50.0 & 56.3 & 50.3 & 53.5 & 52.0 & 52.7 & 50.0 & 50.0 & 51.8 & 50.9 & 70.2 & 50.1 & 50.0 & 100 & \textbf{100} & 54.0 & 100 & 59.3 & 56.8 \\
    GeoAware~\cite{zhang2024telling} & \textbf{96.1} & \textbf{71.0} & 52.1 & 50.0 & 90.6 & 50.3 & 50.6 & 51.4 & \textbf{93.4} & 50.0 & 50.0 & 50.0 & 55.3 & \textbf{80.9} & 50.5 & 50.0 & 100 & 53.0 & 51.0 & 100 & 63.2 & 52.2 \\
    $\chi$\textsubscript{\scalebox{0.8}{BeCoS}}~\cite{wang2025kh} & 76.2 & 57.0 & 95.8 & 63.6 & 66.7 & 78.8 & 72.4 & 97.7 & 82.9 & 72.9 & 63.8 & 96.2 & 81.3 & 60.6 & 93.2 & 80.8 & 100 & 85.2 & 67.2 & 79.0 & 77.3 & 77.3 \\
    Ours\textsubscript{\scalebox{0.8}{BeCoS-PFPascal}} & 90.9 & 50.0 & \textbf{100} & \textbf{100} & \textbf{95.9} & \textbf{90.2} & 88.3 & \textbf{99.2} & 89.1 & \textbf{100} & 93.8 & \textbf{99.5} & \textbf{83.2} & 60.2 & \textbf{96.7} & \textbf{98.2} & \textbf{100} & {95.3} & 93.9 & \textbf{100} & \textbf{86.8} & \textbf{86.8} \\
    Ours\textsubscript{\scalebox{0.8}{BeCoS-Both}} & 90.0 & 50.0 & 93.7 & \textbf{100} & 94.6 & 90.0 & \textbf{88.5} & 98.9 & 89.1 & \textbf{100} & \textbf{94.8} & \textbf{99.5} & 82.4 & 59.2 & \textbf{96.7} & \textbf{98.2} & \textbf{100} & 93.8 & \textbf{95.3} & \textbf{100} & 86.3 & 86.3 \\
    \bottomrule
  \end{tabular}}
\end{table*}

We also compute an overall accuracy $\hat{acc}$ across all categories using \cref{eq: acc_s}, in order to evaluate the inter-class consistency, with
$N$ being the number of images of all categories, $K_i$ being the number of keypoints of category for image $i$, and $y_i \in \{-1,1 \}^{K_i}, y_i^{\text{gt}} \in \{-1,0,1 \}^{K_i}$ denoting the predictions and ground truth of image $i$, respectively. 

\cref{tab:spair-sparse} and \cref{tab:pfpascal-sparse} conclude the quantitative results of left-right prediction on \textsc{SPair-71k} \cite{min2019spair} and \text{PF-Pascal} \cite{ham2016}, respectively. For each dataset, we train our model using \text{BeCoS} \cite{ehm2025beyond} with it for fair comparison, denoted as Ours\textsubscript{\scalebox{0.8}{BeCoS-Spair-71k}} and Ours\textsubscript{\scalebox{0.8}{BeCoS-PF-Pascal}}, respectively. We also train a model using \textsc{BeCoS} \cite{ehm2025beyond} with both datasets, denoted as Ours\textsubscript{\scalebox{0.8}{BeCoS-Both}}. We include qualitative comparison between the most competitive baseline $\chi$\textsubscript{\scalebox{0.8}{BeCoS}} \cite{wang2025kh} and Ours\textsubscript{\scalebox{0.8}{BeCoS-Both}} in \cref{fig:qualitative_comparison}. More qualitative results are in supplementary materials.

\newcommand{\imageheight}{0.07\textheight}

\setlength{\tabcolsep}{0.5pt}
\begin{figure*}[!ht]
    \centering
    \resizebox{\textwidth}{!}{\begin{tabular}{@{}ccccccccc@{}}

\multirow{2}{*}{\rotatebox[origin=c]{90}{$\chi$\textsubscript{\scalebox{0.8}{BeCoS}}~\cite{wang2025kh}}} & \adjustbox{valign=m}
{\includegraphics[height=\imageheight]{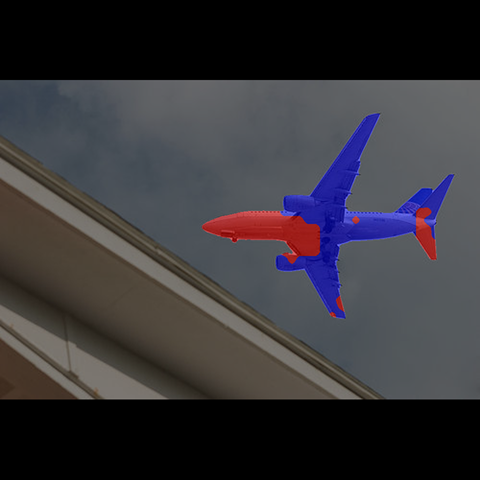}} &
\adjustbox{valign=m}
{\includegraphics[height=\imageheight]{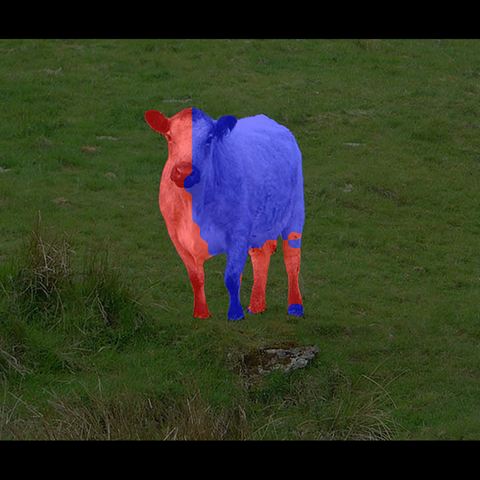}} &
\adjustbox{valign=m}
{\includegraphics[height=\imageheight]{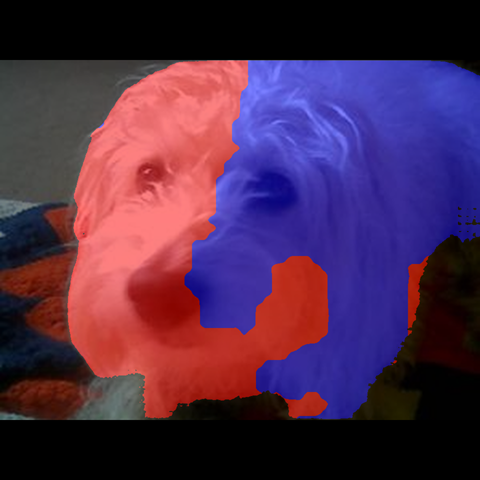}} &
\adjustbox{valign=m}
{\includegraphics[height=\imageheight]{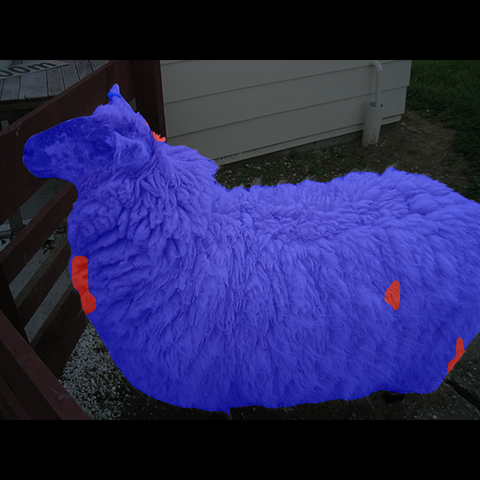}} &
\adjustbox{valign=m}
{\includegraphics[height=\imageheight]{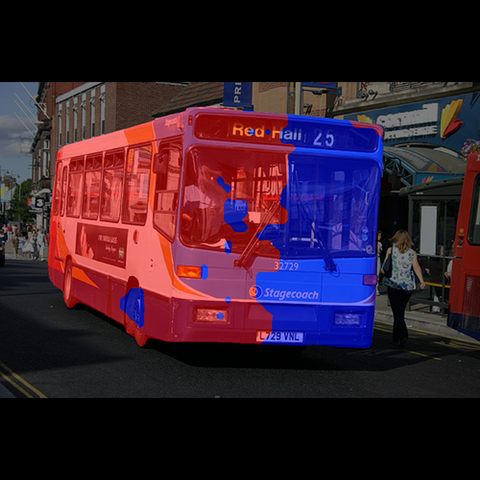}} &
\adjustbox{valign=m}
{\includegraphics[height=\imageheight]{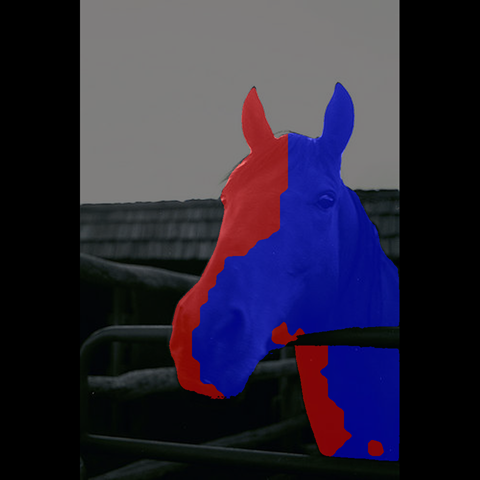}} &
\adjustbox{valign=m}
{\includegraphics[height=\imageheight]{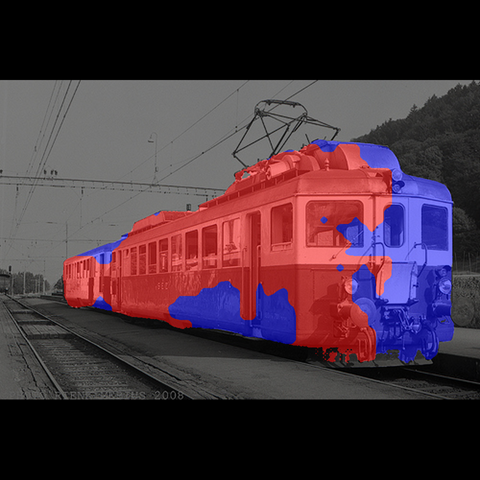}} &
\adjustbox{valign=m}
{\includegraphics[height=\imageheight]{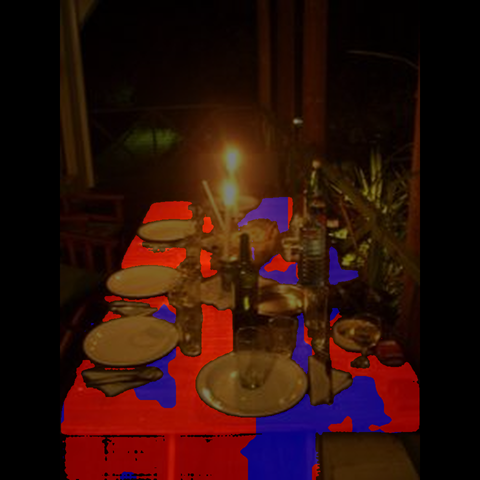}} \\
[1.4em]
 & \adjustbox{valign=m}
{\includegraphics[height=\imageheight]{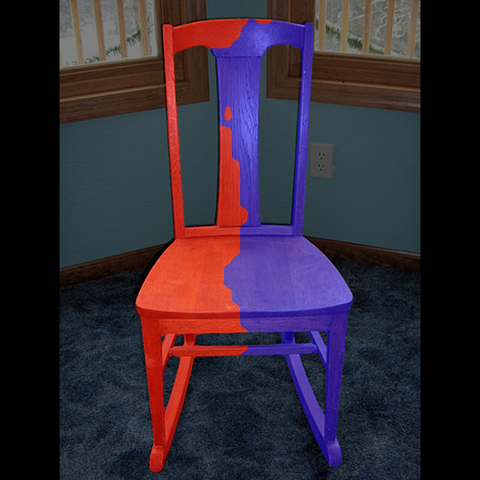}} &
\adjustbox{valign=m}
{\includegraphics[height=\imageheight]{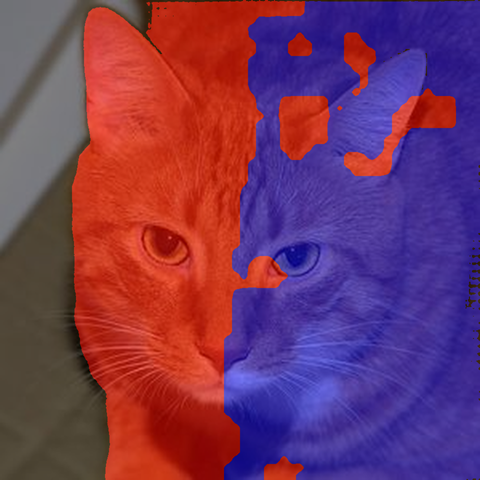}} &
\adjustbox{valign=m}
{\includegraphics[height=\imageheight]{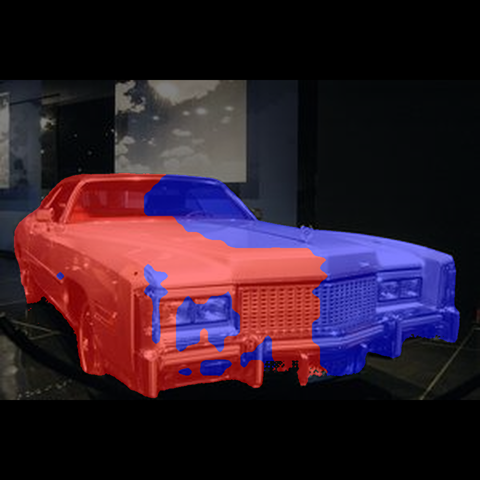}} &
\adjustbox{valign=m}
{\includegraphics[height=\imageheight]{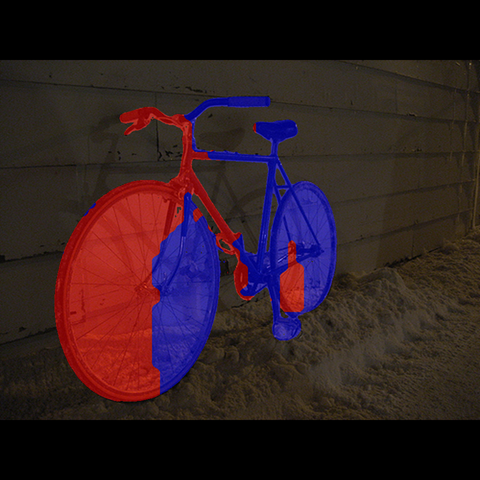}} &
\adjustbox{valign=m}
{\includegraphics[height=\imageheight]{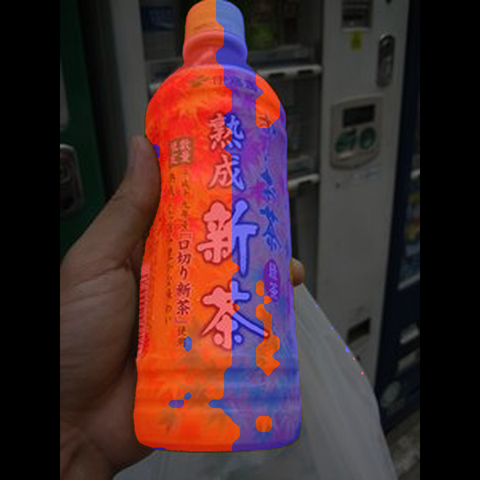}} &
\adjustbox{valign=m}
{\includegraphics[height=\imageheight]{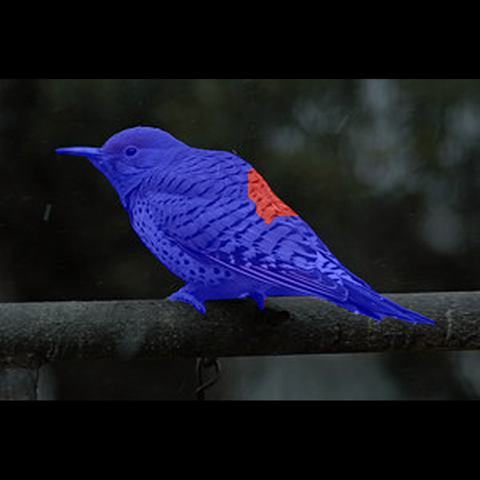}} &
\adjustbox{valign=m}
{\includegraphics[height=\imageheight]{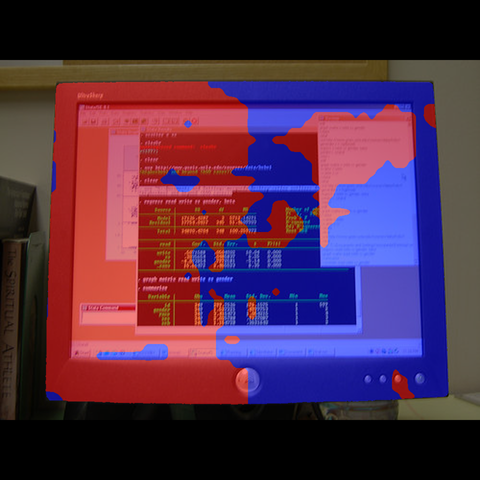}} &
\adjustbox{valign=m}
{\includegraphics[height=\imageheight]{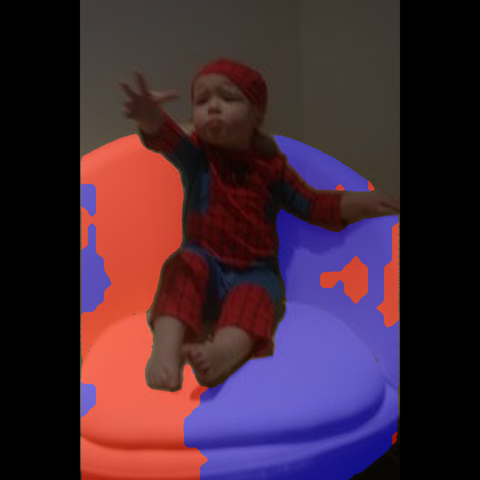}} \\
[1.5em]
\multirow{2}{*}{\rotatebox[origin=c]{90}{Ours\textsubscript{\scalebox{0.8}{BeCoS-Both}}}} & \adjustbox{valign=m}
{\includegraphics[height=\imageheight]{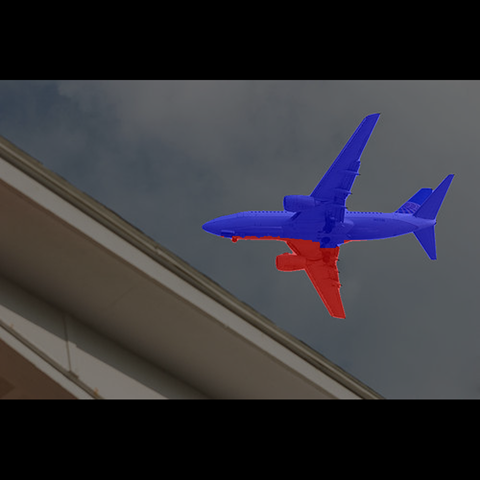}} &
\adjustbox{valign=m}
{\includegraphics[height=\imageheight]{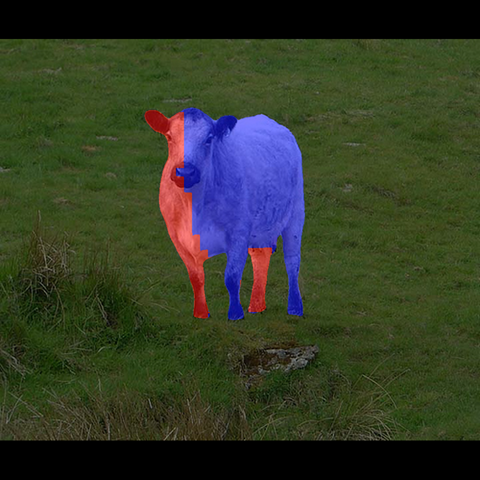}} &
\adjustbox{valign=m}
{\includegraphics[height=\imageheight]{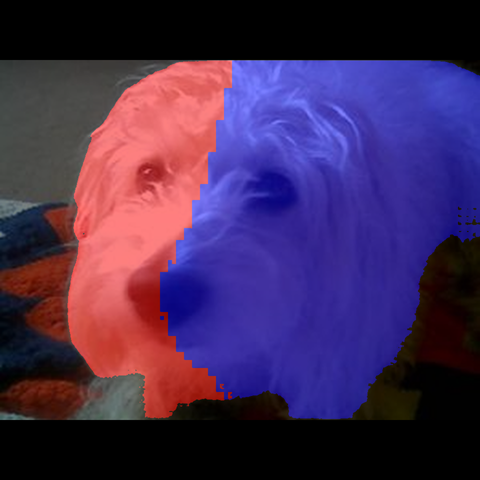}} &
\adjustbox{valign=m}
{\includegraphics[height=\imageheight]{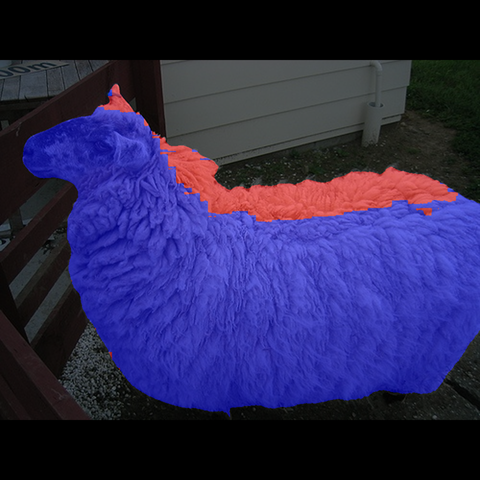}} & 
\adjustbox{valign=m}
{\includegraphics[height=\imageheight]{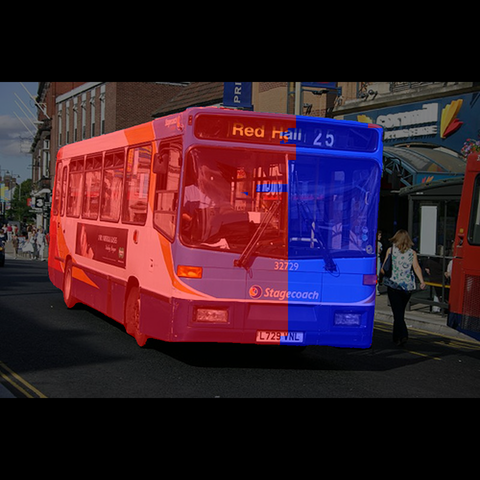}} &
\adjustbox{valign=m}
{\includegraphics[height=\imageheight]{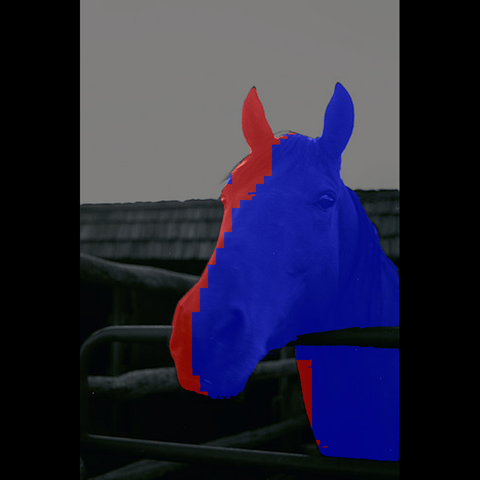}} &
\adjustbox{valign=m}{\includegraphics[height=\imageheight]{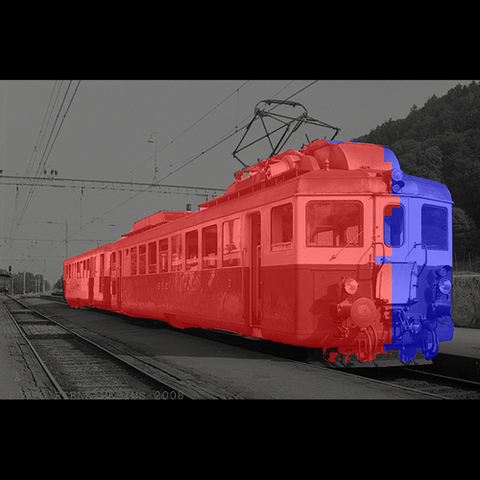}} &
\adjustbox{valign=m}
{\includegraphics[height=\imageheight]{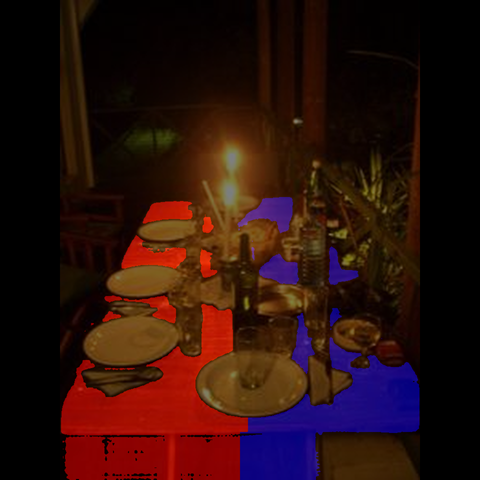}} \\
[1.4em]
 & \adjustbox{valign=m}
{\includegraphics[height=\imageheight]{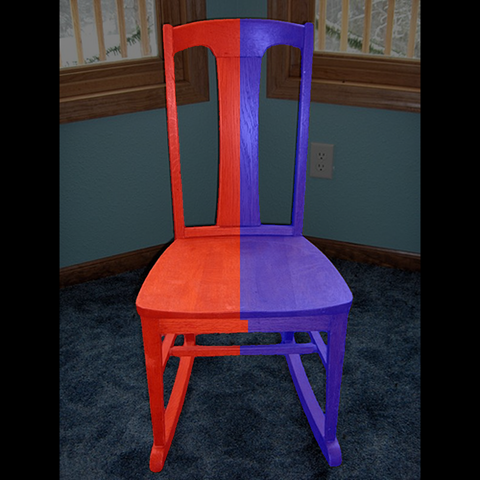}} &
\adjustbox{valign=m}
{\includegraphics[height=\imageheight]{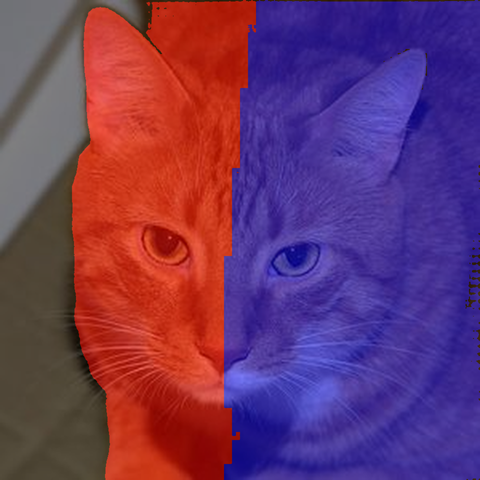}} &
\adjustbox{valign=m}
{\includegraphics[height=\imageheight]{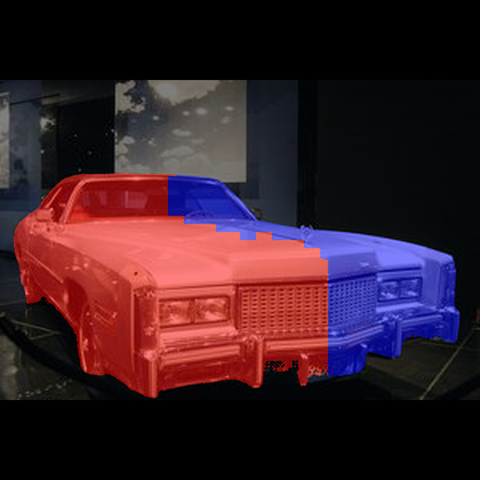}} &
\adjustbox{valign=m}
{\includegraphics[height=\imageheight]{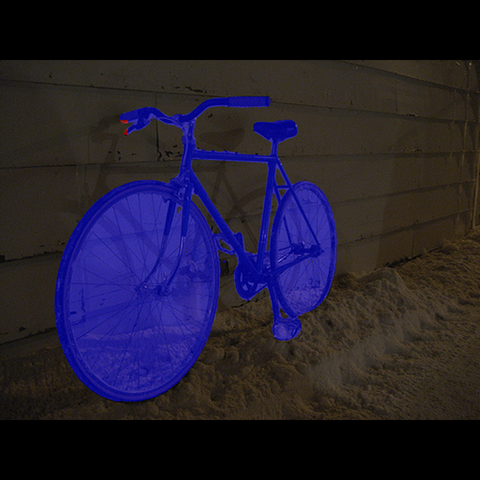}} &
\adjustbox{valign=m}
{\includegraphics[height=\imageheight]{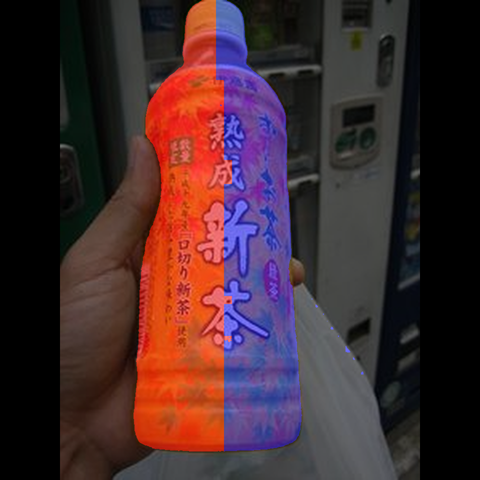}} &
\adjustbox{valign=m}
{\includegraphics[height=\imageheight]{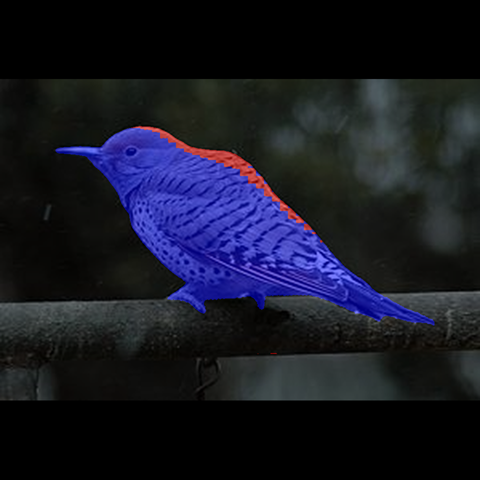}} &
\adjustbox{valign=m}
{\includegraphics[height=\imageheight]{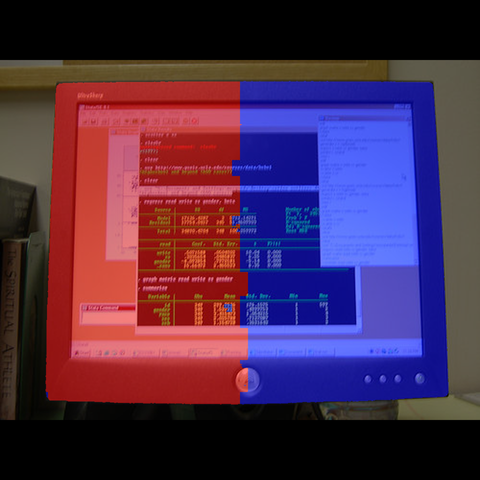}} &
\adjustbox{valign=m}
{\includegraphics[height=\imageheight]{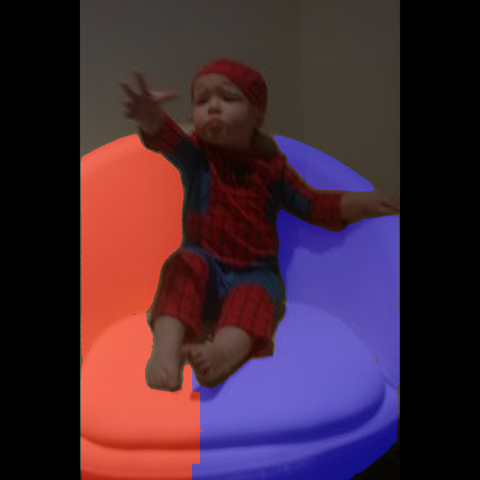}} \\
    \end{tabular}}
    \caption{Qualitative comparisons of dense pixel-level left-right predictions for $\chi$\textsubscript{\scalebox{0.8}{BeCoS}}~\cite{wang2025kh} and Ours\textsubscript{\scalebox{0.8}{BeCoS-Both}} for different object categories sampled randomly from both \textsc{SPair-71k} \cite{min2019spair} and \textsc{PF-Pascal} \cite{ham2016}. Our predictions are significantly more accurate.
    }
    \label{fig:qualitative_comparison}
\end{figure*}

\subsection{Pixel-level Left-Right Prediction on Rendered Images}

For rendered images from test shapes of \textsc{BeCoS} \cite{ehm2025beyond}, for each category, the accuracy $acc$ is defined as in \cref{eq: acc_s}, with
$N$ being the number of images within this category, $y_i \in \{-1, 1\}^{H \times W}$ the pixel-wise prediction, and $y_i^{\text{gt}} \in \{-1, 0, 1\}^{H \times W}$ the ground truth of image $i$. For a pixel $p$ outside the object mask, $y_i^\text{gt}[p] = 0$. 

For evaluating inter-class consistency of the semantic left-right predictions, we also include an overall accuracy $\hat{acc}$ across all categories of rendered images from \textsc{BeCoS} \cite{ehm2025beyond} based on \cref{eq: acc_s}, with $N$ being the number of images of all categories.

\cref{tab:dense} summarises the performance on left-right prediction of all three versions of our model (Ours\textsubscript{\scalebox{0.8}{BeCoS-Spair-71k}}, Ours\textsubscript{\scalebox{0.8}{BeCoS-PF-Pascal}} and Ours\textsubscript{\scalebox{0.8}{BeCoS-Both}}) and baselines, for each category within \textsc{BeCoS} \cite{ehm2025beyond} test dataset. Qualitative results are included in the supplementary materials.

\begin{table}[tbh]
 \setlength{\tabcolsep}{3.4pt}
 \centering
 \caption{Left-Right predictions on rendered images.
 Our method outperforms its competitors by a large margin. Since in this setting the images are rendered from the 3D shape dataset \textsc{BeCoS} \cite{ehm2025beyond}, the performance is similar independent of the choice of in-the-wild image training set.}
  \label{tab:dense}
  \footnotesize
\begin{tabular}{@{}l|ccccccccccc@{}}
    \toprule
    \textbf{Method} & \faTiger & \faBear & \faCat & \faBuck & \faCattle & \faWalking & \faCow & \faGiraffe & \faRhino & $\stackrel{\textbf{AVG}}{acc}$ & $\hat{acc}$ \\
    \midrule
    SD+DINO\cite{zhang2023tale} & 81.8 & 82.3 & 82.8 & 83.5 & 82.9 & 87.5 & 82.1 & \textbf{79.9} & 84.8 & 83.7 & 83.7\\
    SD+DINOv3  & 78.1 & 79.8 & 78.7 & 79.9 & 79.6 & 84.1 & 79.0 & 74.8 & 80.5 & 80.4 & 80.4 \\
    GeoAware\cite{zhang2024telling} & 72.8 & 69.1 & 71.7 & 74.5 & 73.9 & 77.5 & 74.4 & 73.3 & 75.1 & 73.7 & 73.7\\
    $\chi$\textsubscript{\scalebox{0.8}{BeCoS}}\cite{wang2025kh} & 70.8 & 69.8 & 70.0 & 69.4 & 70.3 & 75.2 & 75.2 & 73.9 & 73.1 & 71.4 & 71.4 \\
    \midrule
    Ours\textsubscript{\scalebox{0.8}{BeCoS-SPair71k}} & 83.1 & 89.2 & \textbf{85.2} & 87.1 & 88.0 & \textbf{88.5} & 92.3 & 72.7 & {92.9} & 87.3 & 87.3 \\
    Ours\textsubscript{\scalebox{0.8}{BeCoS-PFPascal}} & 83.3 & 89.1 & \textbf{85.2} & 87.1 & 88.1 & 88.4 & 92.3 & 73.8 & 92.4 & 87.3 & 87.3 \\
    Ours\textsubscript{\scalebox{0.8}{BeCoS-Both}} & \textbf{83.7} & \textbf{89.9} & 84.9 & \textbf{87.3} & \textbf{88.2} & 87.9 & \textbf{92.4} & 75.8 & \textbf{93.7} & \textbf{87.4} & \textbf{87.4} \\
    \bottomrule
  \end{tabular}
\end{table}

\subsection{Generalisation Analysis}
\label{sec: generalisation}

We analyse our trained framework in the following multiple levels to show its strong generalisation ability.

\textbf{Generalisation on 3D information.} The 3D shape dataset \textsc{BeCoS} \cite{ehm2025beyond} that we use as 3D prior only contains human and quadruped animal shapes. As shown in Tab.~\ref{tab:spair-sparse}, \ref{tab:pfpascal-sparse} and \cref{fig:qualitative_comparison}, our Pix2LR also performs well on images containing human-made objects such as aeroplanes, trains, and TVs, despite the fact that no explicit 3D information about these objects is used during training.

\newcommand{\imageheightA}{0.1\textheight}

\setlength{\tabcolsep}{0.5pt}
\begin{figure}[!ht]
    \centering
    \resizebox{\columnwidth}{!}{\begin{tabular}{cccccccc}

\adjustbox{valign=m}
{\includegraphics[height=\imageheightA]{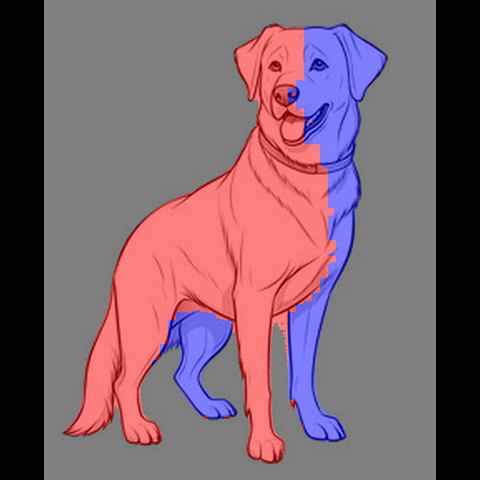}} &
\adjustbox{valign=m}
{\includegraphics[height=\imageheightA]{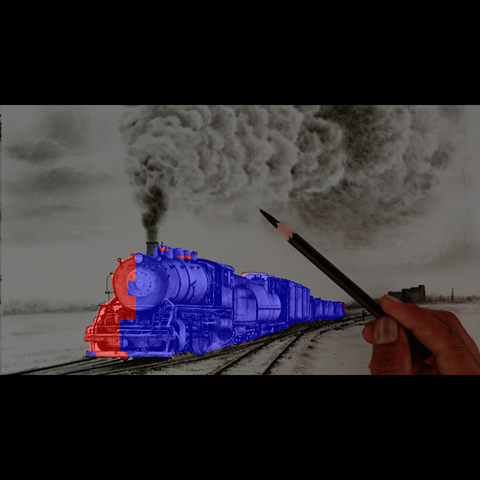}} &
\adjustbox{valign=m}
{\includegraphics[height=\imageheightA]{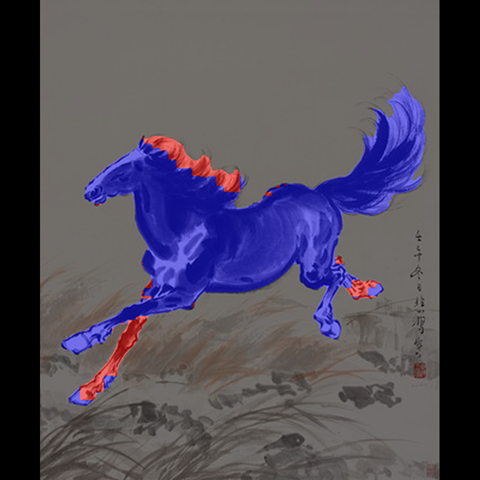}} &
\adjustbox{valign=m}
{\includegraphics[height=\imageheightA]{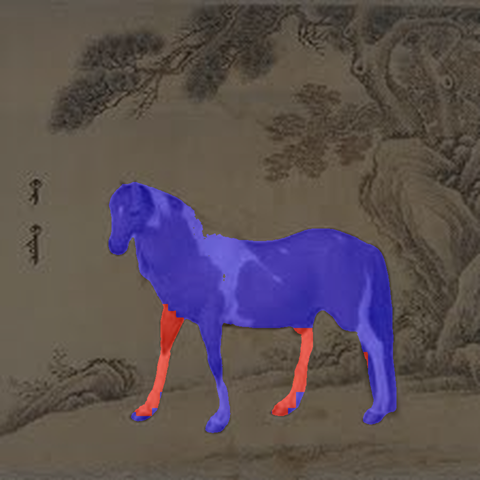}} &
\adjustbox{valign=m}
{\includegraphics[height=\imageheightA]{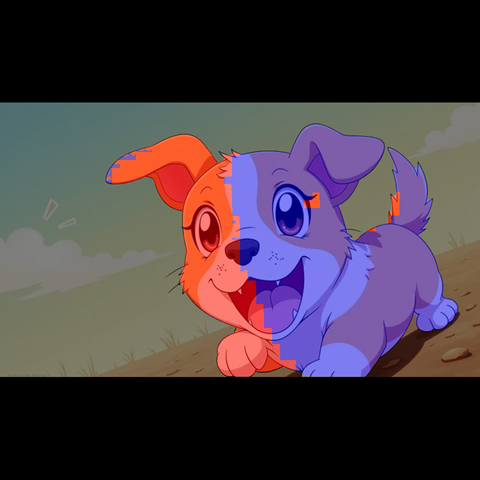}} &
\adjustbox{valign=m}
{\includegraphics[height=\imageheightA]{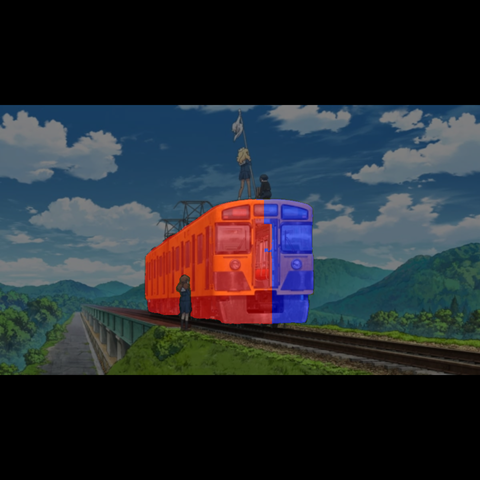}} &
\adjustbox{valign=m}
{\includegraphics[height=\imageheightA]{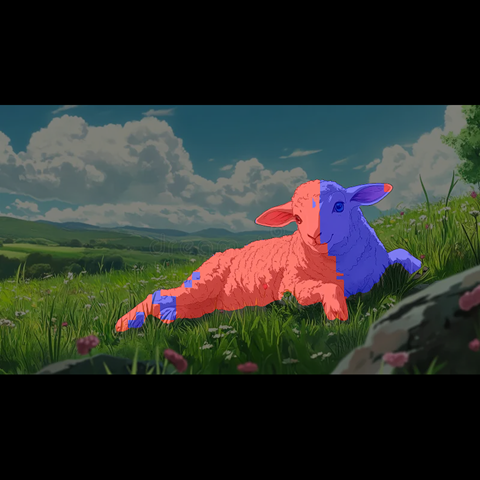}} &
\adjustbox{valign=m}
{\includegraphics[height=\imageheightA]{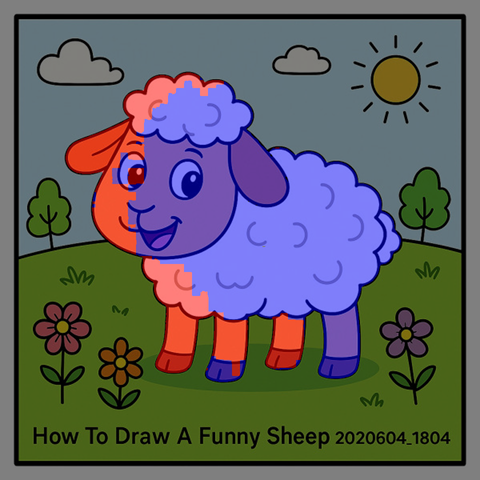}} \\
[2.5em]
\adjustbox{valign=m}
{\includegraphics[height=\imageheightA]{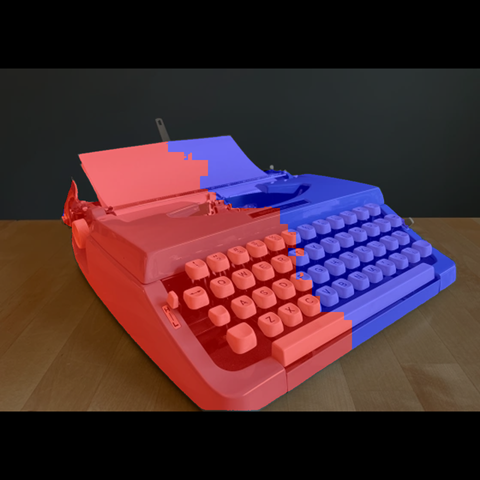}} &
\adjustbox{valign=m}
{\includegraphics[height=\imageheightA]{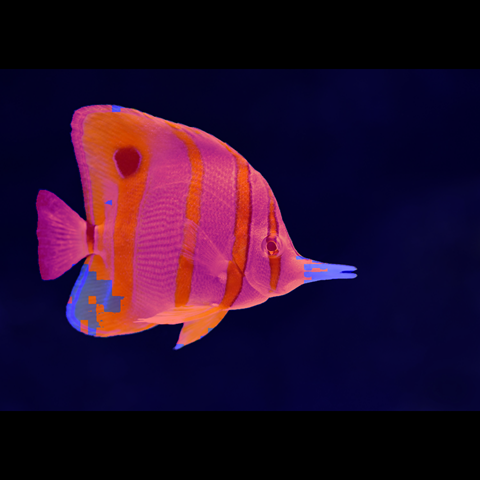}} &
\adjustbox{valign=m}
{\includegraphics[height=\imageheightA]{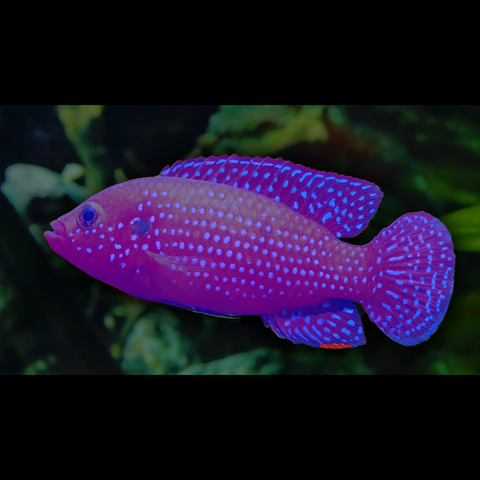}} &
\adjustbox{valign=m}
{\includegraphics[height=\imageheightA]{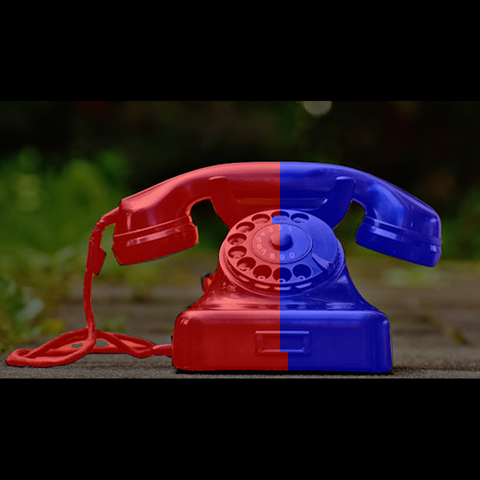}} &
\adjustbox{valign=m}
{\includegraphics[height=\imageheightA]{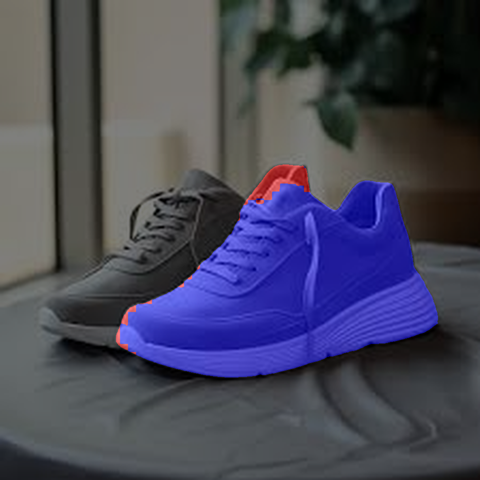}} &
\adjustbox{valign=m}
{\includegraphics[height=\imageheightA]{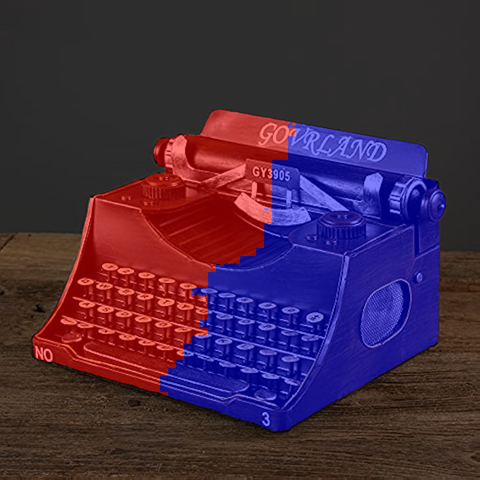}} &
\adjustbox{valign=m}
{\includegraphics[height=\imageheightA]{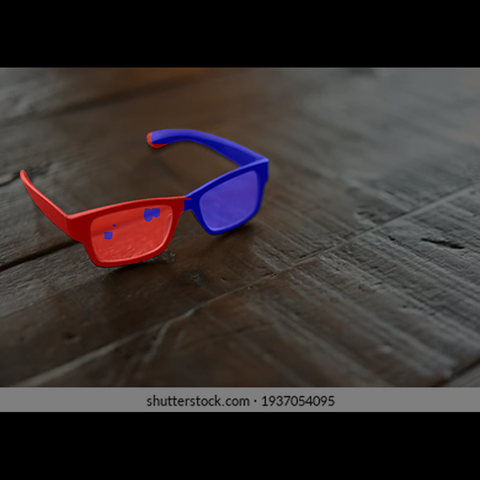}} &
\adjustbox{valign=m}
{\includegraphics[height=\imageheightA]{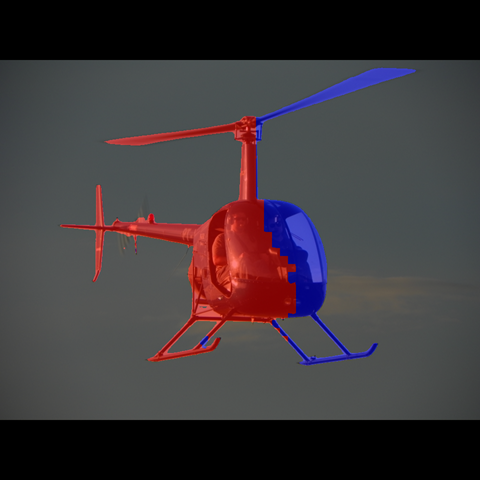}} \\

    \end{tabular}}
    \caption{Zero-shot performance of our method on paintings or anime images (\textit{top}), and on unseen categories neither part of the 3D shape nor the image dataset (\textit{bottom}). }
    \label{fig:style_generalisation}
\end{figure}

\textbf{Generalisation on image geometry variations.} As shown in \cref{fig:qualitative_comparison}, images within \textsc{SPair-71k} \cite{min2019spair} and \textsc{PF-Pascal} \cite{ham2016} contain images of objects from different categories with large pose and viewpoint variations, as well as occlusion and truncation. Yet, our framework can still performs well on most of these cases, which shows generalisation on images with large geometry variations.

\textbf{Generalisation on image style variations.} We further test the performance of our trained PixLR on images with different styles, such as anime or paintings. As shown in the first row of \cref{fig:style_generalisation}, our model can still predict reasonable left-right semantic labels, proves its generalisation ability on image styles.

\textbf{Generalisation on unseen categories.} We evaluate our trained Pix2LR on categories that are neither part of the 3D shape dataset, nor of the image dataset.
Second row in \cref{fig:style_generalisation} shows that the results are reasonable and consistent with the other object categories.

\subsection{Ablation Analysis}
\label{sec: ablation}

We first conduct an ablation analysis on losses by leaving out individual loss terms. Further, we study the relevance of the employed 3D shape dataset used as 3D geometry prior during training. We use the left-right prediction accuracy $\hat{acc}$ of all categories as metrics for either in-the-wild images from  \textsc{SPair-71k} \cite{min2019spair} or rendered images from \textsc{BeCoS} \cite{ehm2025beyond}, for fair and comprehensive evaluation. Results are reported in \cref{tab:ablation}.

\begin{table}
\setlength{\tabcolsep}{7.5pt}
  \centering
  \caption{Ablation analysis when leaving out individual loss terms (\textit{left}), and when training on different 3D shape datasets (\textit{right}). The numbers represent $\hat{acc}$ on keypoints and pixels, respectively. We confirm the necessity of each proposed loss, and conclude that adopting 3D shapes from more categories as in \textsc{BeCoS} \cite{ehm2025beyond} indeed offers richer 3D priors. Note that although models trained without $\mathcal{L}_{\text{img}}$ achieves the same performance to the full model on rendered images, the large margin of performance on more challenging in-the-wild images from \textsc{SPair-71k} \cite{min2019spair} validates its necessity.}
  \label{tab:ablation}
  \resizebox{1.0\columnwidth}{!}{%
  \begin{tabular}{@{}lcccccc@{}}
    \toprule
    & \multicolumn{4}{c}{\textbf{Ablation on losses}} & \multicolumn{2}{c}{\textbf{Ablation on 3D datasets}} \\
    \cmidrule(r){2-5}
    \cmidrule(r){6-7}
    & w/o $\mathcal{L}_{\text{var}}$ &  w/o $\mathcal{L}_{\text{dis}}$ &  w/o $\mathcal{L}_{\text{fif}}$ &  w/o $\mathcal{L}_{\text{img}}$ & \textsc{FAUST} & \textsc{BeCoS} \\
    \midrule
    In-the-wild Images (\textsc{SPair-71k} \cite{min2019spair}) & 55.6 & 56.8 & 60.9 & 73.8 & 78.2 & \textbf{83.8} \\
    Rendered Images (\textsc{BeCoS} \cite{ehm2025beyond}) & 53.9 & 50.3 & 61.7 & \textbf{87.4} & 70.7 & \textbf{87.4} \\
    \bottomrule
  \end{tabular}
  }
\end{table}
\newcommand{\imageheightZ}{0.1\textheight}

\setlength{\tabcolsep}{0.5pt}
\begin{figure}[!ht]
    \centering
    \resizebox{\columnwidth}{!}{\begin{tabular}{cccccccc}
\multicolumn{2}{c}{Multiple Symmetries} & \multicolumn{2}{c}{Unusual Poses} & \multicolumn{2}{c}{Out-of-distribution Views} & \multicolumn{2}{c}{Rare Geometries} \\
\adjustbox{valign=m}
{\includegraphics[height=\imageheightA]{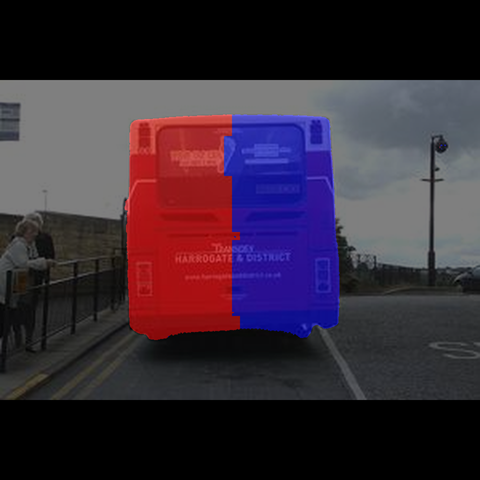}} &
\adjustbox{valign=m}
{\includegraphics[height=\imageheightA]{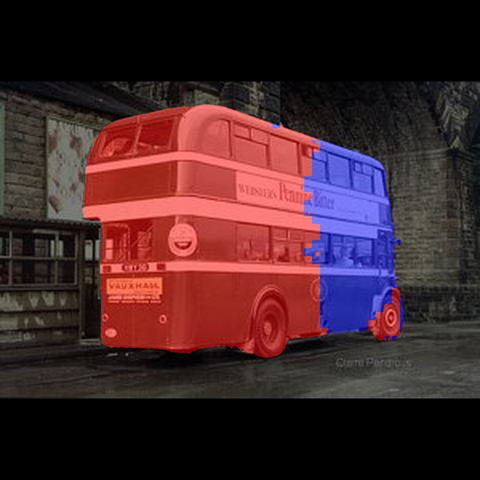}} &
\adjustbox{valign=m}
{\includegraphics[height=\imageheightA]{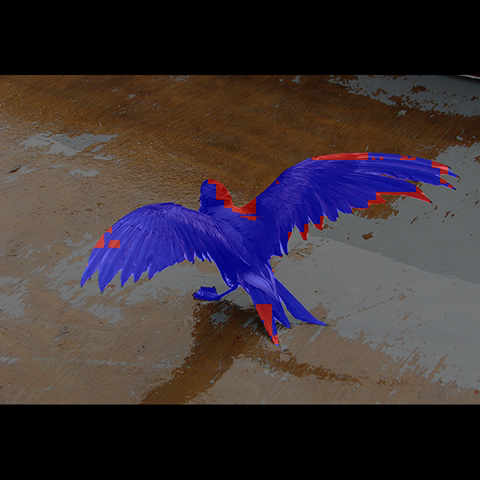}} &
\adjustbox{valign=m}
{\includegraphics[height=\imageheightA]{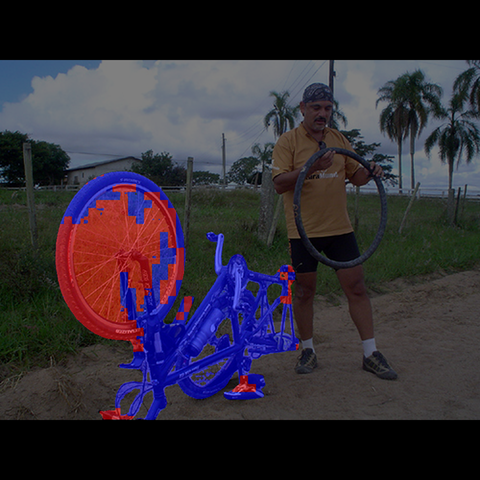}} &
\adjustbox{valign=m}
{\includegraphics[height=\imageheightA]{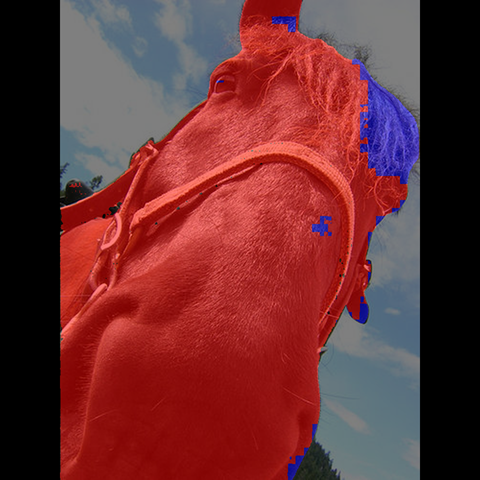}} &
\adjustbox{valign=m}
{\includegraphics[height=\imageheightA]{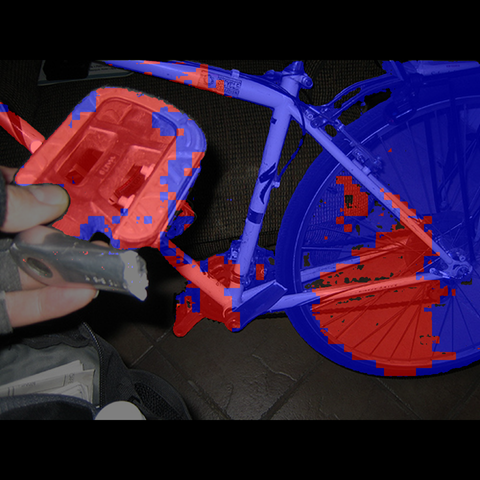}} &
\adjustbox{valign=m}
{\includegraphics[height=\imageheightA]{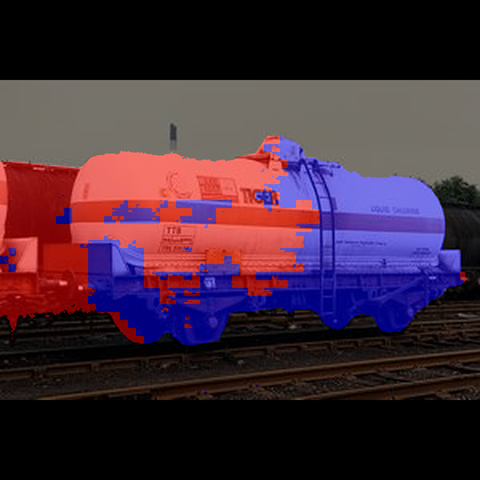}} &
\adjustbox{valign=m}
{\includegraphics[height=\imageheightA]{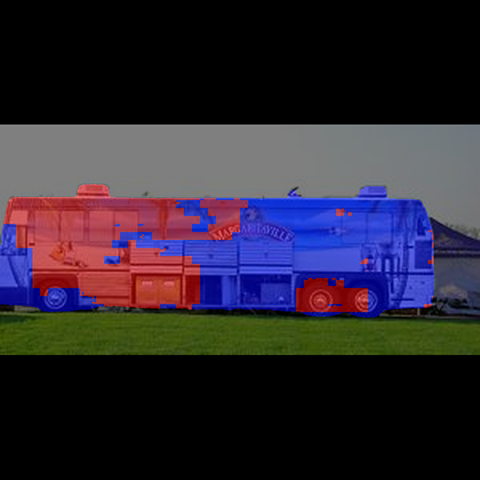}} \\
    \end{tabular}}
    \caption{We identify a few challenges for method leading to wrong predictions:
    (i) multiple symmetries,
    (ii) out-of-distribution views , (iii) unusual poses, (iv) rare geometries.}
    \label{fig:limitations}
\end{figure}
\section{Discussion \& Limitations}

Overall, our Pix2LR shows superiority compared to other state-of-the-art methods for semantic left-right understanding of images across a broad range of categories. Nevertheless, there is room for improvements. First, as shown in \cref{fig:limitations}, various hard cases may lead to inconsistent predictions.
Future works that combine our framework with data augmentation, or leverage larger-scale 3D shape datasets with more versatile categories as geometric prior, may alleviate these difficulties. Besides, for some object categories which are rather `thin' along the left-right axis, such as bicycle, it is hard to predict accurate left-right values of fine structures, such as handlebars or pedals, see \cref{fig:qualitative_comparison}. We believe that the addition of attention blocks for an instance-specific information exchange is an interesting direction. 
Finally, our current formalism only considers left-right symmetry, which is for example violated for rotationally symmetric objects or objects with more than two symmetries (\eg, bottles or tables, in \cref{fig:qualitative_comparison}). Empirically, our methods produces consistent results in these cases -- we observe that this ambiguity is implicitly resolved  by interpreting the semantically `frontal part' of the object to be facing the camera, so that
visible pixels of respective objects are split more or less evenly into left and right regions. An interesting direction for future work is to consider more expressive symmetry representations.
\section{Conclusion}
\label{sec: conclusion}
The task of pixel-wise semantic left-right prediction of in-the-wild images is very challenging and under-explored. Difficulties include the
absence of 3D information in single-view images, occlusion, as well as variations in object poses, viewpoints, texture, or geometry. In this work we propose the first unsupervised framework to explicitly detect left-right parts for objects within in-the-wild images from a broad set of categories. We tackle this by leveraging the recent medium-scale 3D shape dataset \textsc{BeCoS} \cite{ehm2025beyond} to serve as a powerful geometric 3D prior, and then combine this with diverse in-the-wild images across different categories to train a left-right prediction neural network in an unsupervised manner.
Our experiments qualitatively and quantitatively demonstrate the effectiveness of our proposed framework. 
We believe our work may be beneficial for multiple downstream applications, such as resolving left-right ambiguity in pose estimation, keypoint detection and part segmentation, or enforcing left-right consistency for image generation and editing.

\newpage
\section*{Acknowledgements}
This work is supported by the ERC starting grant no. 101160648 (Harmony). The authors gratefully acknowledge the access to the Marvin cluster of the University of Bonn.

\bibliographystyle{splncs04}
\bibliography{main}

\end{document}